\documentclass[pdflatex,sn-mathphys-num]{sn-jnl}% Math and Physical Sciences Numbered Reference Style 
\usepackage{graphicx}%
\usepackage{multirow}%
\usepackage{amsmath,amssymb,amsfonts}%
\usepackage{amsthm}%
\usepackage{mathrsfs}%
\usepackage[title]{appendix}%
\usepackage{xcolor}%
\usepackage{textcomp}%
\usepackage{manyfoot}%
\usepackage{booktabs}%
\usepackage{algorithm}%
\usepackage{algorithmicx}%
\usepackage{algpseudocode}%
\usepackage{listings}%
\usepackage{float}
\usepackage{booktabs} 

\theoremstyle{thmstyleone}%
\theoremstyle{thmstyletwo}%

\theoremstyle{thmstylethree}%

\begin{document}

\title[Article Title]{A Cognitive-Mechanistic Human Reliability Analysis Framework: A Nuclear Power Plant Case Study}

\author[1,2]{ \sur{Xingyu Xiao}}
\author[3]{ \sur{Peng Chen}}

\author[1,2]{\sur{Jiejuan Tong}}
\author[4]{ \sur{Shunshun Liu}}
\author[1,2]{\sur{Hongru Zhao}}
\author[1,2]{\sur{Jun Zhao}}
\author[1]{\sur{Qianqian Jia}}
\author*[1]{\sur{Jingang Liang}}
\author[1]{\sur{Haitao Wang}}

\affil[1]{\orgdiv{Institute of Nuclear and New Energy Technology}, \orgname{Tsinghua University}, \orgaddress{ \city{Beijing}, \postcode{100084}, \country{China}}}    

\affil[2]{\orgdiv{National Key Laboratory of Human Factors Engineering}, \orgname{China Astronaut Research and Training Center}, \orgaddress{ \city{Beijing}, \postcode{100094}, \country{China}}}  

\affil[3]{\orgdiv{Software Institute}, \orgname{Chinese Academy of Sciences}, \orgaddress{ \city{Beijing}, \postcode{100086},  \country{China}}}

\affil[4]{\orgdiv{CNNC Shandong Nuclear Energy Co., Ltd}, \orgaddress{ \city{Qingdao}, \postcode{266071}, \state{Shandong} \country{China}}}

\abstract{
Traditional human reliability analysis (HRA) methods, such as IDHEAS-ECA, rely on expert judgment and empirical rules that often overlook the cognitive underpinnings of human error. Moreover, conducting human-in-the-loop experiments for advanced nuclear power plants is increasingly impractical due to novel interfaces and limited operational data. This study proposes a cognitive-mechanistic framework (COGMIF) that enhances the IDHEAS-ECA methodology by integrating an ACT-R-based human digital twin (HDT) with TimeGAN-augmented simulation. The ACT-R model simulates operator cognition, including memory retrieval, goal-directed procedural reasoning, and perceptual-motor execution—under high-fidelity scenarios derived from a high-temperature gas-cooled reactor (HTGR) simulator. To overcome the resource constraints of large-scale cognitive modeling, TimeGAN is trained on ACT-R-generated time-series data to produce high-fidelity synthetic operator behavior datasets. These simulations are then used to drive IDHEAS-ECA assessments, enabling scalable, mechanism-informed estimation of human error probabilities (HEPs). Comparative analyses with SPAR-H and sensitivity assessments demonstrate the robustness and practical advantages of the proposed COGMIF. Finally, procedural features are mapped onto a Bayesian network to quantify the influence of contributing factors, revealing key drivers of operational risk. This work offers a credible and computationally efficient pathway to integrate cognitive theory into industrial HRA practices.}

\keywords{ Human Reliability, Human Digital Twins, IDHEAS-ECA, TimeGAN, Bayesian}

\maketitle

\section{Introduction}\label{sec1}

Human reliability analysis (HRA) plays a pivotal role in the safety assessment of complex socio-technical systems, particularly in high-risk domains such as nuclear power generation \cite{xiao2025dynamic}. As a fundamental component of probabilistic risk assessment (PRA), HRA aims to estimate the likelihood of human error under specific operational contexts, thereby supporting risk-informed decision-making and the design of resilient safety systems. Over the past decades, a range of structured methodologies, such as the standardized plant analysis risk-human reliability analysis (SPAR-H) \cite{gertman2005spar}, the technique for human error rate prediction (THERP) \cite{boring2012fifty}, and more recently, the integrated human event analysis system for event and condition assessment (IDHEAS-ECA) \cite{xing2020integrated}, have been developed to quantify human error probabilities (HEPs). While these frameworks offer operational utility, they are primarily grounded in expert judgment, predefined performance shaping factors (PSFs), and empirically derived databases, often lacking a mechanistic understanding of the cognitive processes that drive operator actions and errors.

Furthermore, traditional HRA approaches are highly dependent on two major data sources: (1) retrospective analysis of operational events, and (2) human-in-the-loop (HITL) simulation experiments conducted in controlled environments. Representative efforts in the latter category include the human reliability experimental (HuREx) system \cite{jung2020hurex} and the scenario authoring, characterization, and debriefing application (SACADA) platform \cite{chang2014sacada}, both of which aim to systematically capture operator behavior under realistic but simulated conditions. While valuable, HITL-based data acquisition is inherently resource-intensive, time-consuming, and difficult to scale, particularly in the context of next-generation nuclear power plants. In these emerging systems, novel human–machine interfaces \cite{zhang2002future}, automation-intensive control architectures \cite{locatelli2014small}, and limited operational history \cite{jia2014control} pose formidable challenges to the applicability and scalability of existing HRA paradigms \cite{miao2023comprehensive}.

As operator–system interaction becomes increasingly complex, there is a growing imperative to develop mechanistically grounded and computationally scalable HRA methodologies that can systematically incorporate cognitive, procedural, and contextual dynamics into risk-informed analyses. Addressing this need requires novel frameworks that can both simulate realistic operator behavior and generate sufficient data volume to support probabilistic modeling without relying exclusively on costly human-in-the-loop experiments.

To address these limitations, this study proposes a cognitive-mechanistic human reliability analysis framework (COGMIF) that enhances the IDHEAS-ECA methodology through the integration of ACT-R-based cognitive simulation and GAN-augmented behavioral data generation. The framework is grounded in the adaptive control of thought-rational (ACT-R) cognitive architecture, which mechanistically models human information processing, including memory retrieval, goal-directed procedural reasoning, and perceptual-motor execution. Simulated experiments were conducted on a high-temperature gas-cooled reactor (HTGR) simulator under representative operational scenarios, including multi-parameter queries with interface switching, parameter searches without interface transitions, operations involving three-phase alternating current (AC) systems, and discrete button-based interactions.

To improve the scalability of this cognitively grounded approach, a two-stage generative model, TimeGAN, is employed. Trained on time-series data produced by ACT-R simulations, TimeGAN generates high-fidelity synthetic operator behavior sequences that preserve temporal dependencies and behavioral plausibility. These augmented datasets are subsequently utilized as inputs to the IDHEAS-ECA framework, enabling systematic and scalable estimation of human error probabilities (HEPs) based on mechanistically derived operator behaviors. Finally, key procedural and cognitive variables are mapped onto a Bayesian network constructed using the synthesized HRA data, allowing for influence strength quantification and sensitivity analysis to identify critical contributors to human error. This integrated framework offers a novel pathway for enhancing the explanatory depth, realism, and scalability of human reliability assessments in complex nuclear operational environments.

The main contributions of this work are summarized as follows:

\begin{itemize} 
\item Development of a cognitively grounded HRA framework COGMIF that integrates the ACT-R cognitive architecture with the IDHEAS-ECA methodology, enabling mechanistic modeling of operator behavior—including memory retrieval, goal-driven reasoning, and motor execution—under representative nuclear power plant scenarios.

\item Introduction of a scalable behavioral data augmentation approach by training TimeGAN on ACT-R-generated time-series, allowing for the generation of high-fidelity synthetic operator behavior datasets that preserve temporal structure and support large-scale human error estimation without reliance on costly human-in-the-loop experiments. 
\item Construction of a Bayesian network informed by cognitive and procedural features derived from the simulated data, facilitating influence strength quantification and sensitivity analysis to identify dominant contributors to human error, thereby enhancing both the interpretability and predictive capability of the proposed framework COGMIF. \end{itemize}

This paper is structured as follows: Section \ref{Literature Review} presents the literature review, Section \ref{Methodology} outlines the methodology, Section \ref{Results and Evaluation} discusses the results and evaluation, and Section \ref{Conclusion and Discussion} concludes with a discussion.

\section{Literature Review}\label{Literature Review}

\subsection{Human Reliability Analysis}

Human reliability analysis (HRA) is a systematic approach used to assess and model human performance and its impact on the reliability of complex systems. HRA seeks to identify potential human errors, quantify their likelihood, and evaluate their consequences, particularly in high-risk domains such as aviation, nuclear power, healthcare, and military operations. The primary goal of HRA is to enhance system safety and performance by addressing the cognitive, behavioral, and environmental factors that contribute to human errors.

Several methods have been developed within the HRA field to model human errors and assess human reliability, including technique for human error rate prediction (THERP) \cite{swain1964therp}, human cognitive reliability (HCR) \cite{hannaman1985model}, SPAR-H \cite{gertman2005spar}, cognitive reliability and error analysis method (CREAM) \cite{hollnagel1998cognitive}, and a technique for human error analysis (ATHEANA) \cite{cooper1996technique}, among others. Currently, HRA has evolved into its third generation. The key characteristics of third-generation HRA are its dynamic nature and the incorporation of cognitive mechanisms into interpretable algorithms. One of the most prominent algorithms is the Integrated Decision Tree Human Event Analysis System for Event and Causal Analysis (IDHEAS-ECA), developed by Xing et al.\cite{xing2020integrated} at the U.S. Nuclear Regulatory Commission (NRC). This method stands out due to its authoritative and up-to-date data sources, IDHEAS-DATA \cite{xing2021draft}, which includes the SACADA database \cite{chang2014sacada} and the publicly available data and various case studies. Moreover, IDHEAS-ECA is a dynamic approach that integrates macro-cognitive aspects, considering five levels: detection, understanding, decision-making, action execution, and inter-team coordination. As illustrated in Figure \ref{IDHEAS}, IDHEAS-ECA consists of a total of eight steps \cite{xing2020integrated}. Additionally, Xiao et al. \cite{xiao2024krail} have implemented a semi-automated approach for determining Base HEPs using Large Language Models (LLMs) and knowledge graphs, but this method lacks a comprehensive calculation for the overall HEP. 

\begin{figure}[h]
\centering
\includegraphics[width=1.0\textwidth]{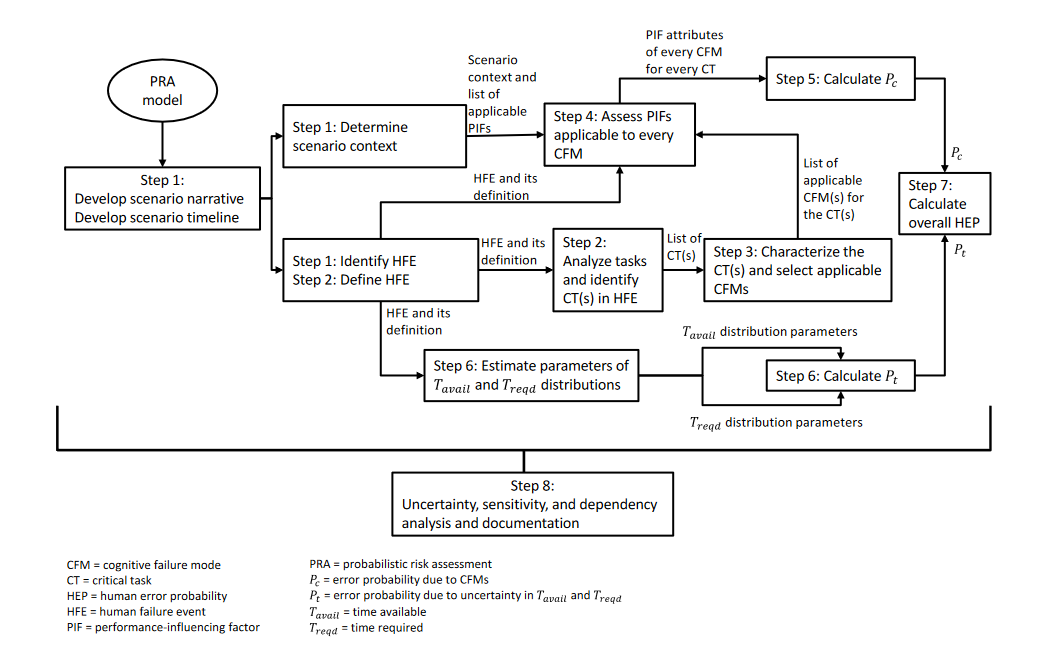}
\caption{IDHEAS-ECA HRA Process \cite{xing2020integrated}}\label{IDHEAS}
\end{figure}

While IDHEAS-ECA provides a structured and widely accepted framework for human error probability estimation, it decomposes human error into two key components: the probability of cognitive failure ($P_{c}$) and the probability of time-based failure ($P_{t}$) \cite{xing2020integrated}. Although cognitive failure is assessed through performance shaping factors and scenario conditions, the estimation of $P_{t}$ typically relies on expert judgment or assumed task duration thresholds. These assumptions are often made without detailed consideration of task-specific cognitive or motor dynamics, resulting in uncertainty and limited generalizability.

To address this limitation, we propose leveraging the ACT-R cognitive architecture to simulate task execution times based on the underlying cognitive and procedural structure of the operator's actions. ACT-R enables explicit modeling of the temporal characteristics of human behavior, including memory retrieval latency, goal formulation delays, and motor execution times, under scenario-specific constraints. By integrating ACT-R-derived time estimates into the IDHEAS-ECA framework, the proposed method enables a more mechanistically justified and data-driven approach to estimating $P_{t}$, thereby reducing reliance on expert heuristics and enhancing the precision of time-sensitive human error analysis.

\subsection{Existing Approaches in HRA Data Collection}

Traditional human reliability analysis (HRA) data collection methodologies primarily rely on retrospective analysis of historical incident reports from nuclear power plants conducted by domain experts. However, this approach faces significant limitations when applied to advanced modular reactor systems. The operational experience of next-generation reactors remains insufficient to establish comprehensive accident case databases. Moreover, fundamental design changes in main control room interfaces and reactor configurations create substantial differences from conventional nuclear plants, rendering HRA data extrapolated from legacy reactor types technically inappropriate.

Contemporary HRA data collection frameworks, notably the human reliability experimental (HuRex) system, the scenario authoring, characterization, and debriefing application (SACADA) approach, and the simplified human error experimental program (SHEEP) attempt to address these challenges. As illustrated in Table \ref{compare}, the HuRex \cite{jung2020hurex} methodology implements a four-phase process: (1) preparation, (2) qualitative data collection, (3) quantitative data analysis, and (4) data reporting. While its predefined taxonomy of 23 task types and error modes enables systematic documentation of human behavior deviations, the process demands extensive expert involvement in reviewing simulator recordings and log analysis. This labor-intensive nature hinders practical engineering applications despite its compatibility with digital control room environments.

The SACADA methodology \cite{chang2014sacada} demonstrates superior engineering viability by integrating data collection into routine operator training programs. This approach automatically archives computer interaction data during pre-configured simulator scenarios, eliminating additional training burdens while generating empirically grounded HRA datasets. Nevertheless, two critical limitations persist: First, residual dependence on expert judgment introduces subjectivity in performance evaluation criteria, particularly in threshold determination for unsatisfactory operator actions. Second, the collected data requires supplementary contextual analysis to fully characterize cognitive processes behind observed behaviors \cite{park2022framework}.

Both methodologies fail to adequately resolve the fundamental challenges posed by next-generation reactor systems: insufficient operational experience, evolving human-system interfaces, and the demand for resource-efficient data collection paradigms. These limitations highlight the necessity for innovative approaches that minimize expert dependency while ensuring a comprehensive capture of context-specific human reliability parameters in advanced nuclear environments.

% The SHEEP employs a cost-effective, flexibly configurable rancor microworld simulator as a substitute for a full-range simulator, thereby lowering the experimental threshold. Participants with basic nuclear engineering training (i.e., students) were incorporated to supplement data from professional operators, effectively expanding the sample size. Through empirical investigation, the study validated performance differences between students and professional operators and explored the potential of simplified simulator data to support comprehensive research. However, several issues remain. First, inherent differences exist between students and operators—for instance, in determining which procedures to implement during emergency phases. Second, the impact of interface characteristics on human reliability analysis (HRA) has been overlooked. Previous studies have demonstrated that interface design significantly influences HRA \cite{rasmussen1989coping}. Particularly in the context of high-Temperature gas-cooled reactor (HTGR) modular systems, where a single operator may control multiple units. Consequently, the simplified rancor microworld simulator may not adequately replicate the complexities of such environments, and the reliance on human resources continues to pose a challenge \cite{park2022framework}.

\begin{table}[h]
\caption{Comparison of Data Collection Approaches: HuRex vs. SACADA}\label{compare}%

\begin{tabular}{p{0.15\linewidth} p{0.45\linewidth} p{0.45\linewidth}}

\toprule

\textbf{Metrix}  & \textbf{HuRex} & \textbf{SACADA} \\
\midrule
\textbf{Data Collection Process} & 
Four stages are implemented: (1) Preparation, (2) Qualitative Data Collection, (3) Quantitative Data Analysis, and (4) Data Reporting.
& 
During each simulator training session, instructors use the existing simulator to pre-set training scenarios and expected operational steps. Operator computer interaction data is collected throughout the exercise and automatically archived into a database. \\[1ex]
\midrule
\textbf{Advantages} & 
Predefined 23 task types and corresponding error modes enable systematic capture of diverse human behavior deviations, thereby reducing the omission of critical information. Moreover, the method incorporates digital control room factors, and its task categorization and data extraction processes are well suited for digital interface environments, supporting human factors data needs in next-generation digital nuclear power plants.
& 
The primary advantage is the seamless integration of human factors data collection into routine operator training, which does not impose any additional training burden. This approach continuously generates valuable data that provides extensive empirical support for quantitative HRA models. \\[1ex]
\midrule
\textbf{Disadvantages} & 
The data collection and analysis process is complex and time-consuming. It requires experienced experts to review extensive simulator recordings and logs to identify and classify unsafe behaviors—a process that remains partly subjective despite assistance from video and data analysis tools.
& 
Consistency and subjectivity issues persist; although SACADA uses clearly defined operational evaluation criteria, variations in instructor scoring thresholds (e.g., determining when performance is “unsatisfactory”) may lead to discrepancies. \\

\botrule
\end{tabular}
\end{table}

\subsection{Advancements in Mechanistic Simulation}

Modeling the cognitive mechanisms of nuclear power plant (NPP) operators has advanced significantly in the past decade, driven by the need for realistic human factors engineering (HFE) simulations to support safety and design. Three notable cognitive architectures, the integrated decision-making and action control (IDAC) model \cite{chang2007cognitive}, the queueing network-Model Human Processor (QN-MHP) \cite{liu2006queueing}, and the adaptive control of thought-rational (ACT-R) \cite{anderson1997act}, have been applied to capture operator decision-making, behavior, and error dynamics in NPP settings. Each offers distinct strengths and limitations in simulating operator cognition and decision processes under complex scenarios. 

IDAC is a cognitively inspired framework originally developed to probabilistically simulate control room crew responses during NPP accidents \cite{andersson2017idac}. It integrates a heuristic cognitive engine and reasoning module to represent how operators perceive information, assess situations, set goals, and execute actions in a team context \cite{azarkhil2025team}. A key advantage of IDAC is its comprehensive scope: it explicitly models individual and crew behavior (including communication and coordination) \cite{xiao2024emergency}, and incorporates performance-influencing factors (e.g. stress, experience) and even emotional states into the decision-making rules. Coupled with a plant simulation (e.g. the Accident Dynamics Simulator) \cite{coyne2014nuclear}, IDAC enables dynamic PRA/HRA by automatic, time-stepped simulation of human–system interactions in accident scenarios. However, IDAC’s breadth comes with challenges. Its cognitive models are relatively coarse-grained (relying on generic rules and heuristics) and require extensive knowledge of engineering to represent plant procedures and operator expertise. Validation studies indicate that IDAC may need calibration to faithfully mimic specific operator populations \cite{chang2007cognitive}, for example, adjustments were required to model novice operator behavior versus expert. In practice, the IDAC framework can be computationally intensive when integrated into large Monte Carlo simulations, and its proprietary complexity means it is primarily applied by specialized teams. While highly useful for scenario-based risk analysis, IDAC is less oriented toward explaining fine-grained cognitive strategies or neural plausibility, focusing instead on macro-level human reliability outcomes.

The QN-MHP architecture takes an engineering approach to human cognition, merging the classic model human processor (MHP) with a queueing network formulation to simulate multitask performance \cite{liu2006queueing}. In QN-MHP, an operator’s perceptual, cognitive, and motor processes are represented as a network of servers that process information with certain service time distributions, enabling a mathematical mapping of concurrent activities and mental workload. This approach has been adapted to the nuclear domain to predict how interface complexity and task load impact operator behavior. For instance, Chen and Tong et al. applied QN-MHP to model a single operator monitoring multiple reactor modules \cite{chen2024influence}; the QN-MHP simulation quantified the cognitive load and error likelihood under different human–machine interface (HMI) designs. However, QN-MHP’s abstract, queuing-based nature means it emphasizes timing and capacity over detailed cognitive strategies. It does not explicitly model knowledge, learning, or high-level decision rules, factors that can be pivotal in novel or diagnosis-heavy scenarios. Interpretability is generally high in terms of workload and processing delays, but lower regarding why an operator takes a given action (since cognitive processes are aggregated into service nodes) \cite{wu2008development}. In addition, while computationally lighter than fully featured cognitive simulations, QN-MHP requires careful task analysis and calibration of server parameters (e.g. service time distributions for mental operators) to reflect nuclear domain tasks. Its reliance on empirical tuning might limit generality across contexts without new data \cite{tsimhoni2003modeling}.

ACT-R is one of the most prominent cognitive architectures in psychology and has increasingly been employed for high-fidelity simulation of operator cognition in process control domains \cite{lebiere2019error}. It consists of modular components for perception, memory, decision-making, and motor functions that interact to produce human-like behavior \cite{taatgen2006modeling}. ACT-R’s cognitive mechanisms (e.g. activation-based memory retrieval, production rule learning) are grounded in a broad base of psychological evidence, enabling it to not only predict task performance times but also mimic errors, strategy shifts, and workload effects as observed in human subjects. Indeed, studies have noted that ACT-R, while “one of the most thorough systems” in scope, is a “relatively complex and hefty architecture” not always ideal for quick exploratory studies \cite{cao2024computational}. Computationally, ACT-R models run in cognitive time scales and may require acceleration or simplification to explore very large state spaces in PRA contexts. Despite these challenges, ACT-R’s strengths – high cognitive fidelity, neural plausibility, and strong empirical support – make it a powerful tool for mechanistic simulation of operator behavior, especially when the goal is to understand subtle cognitive phenomena (e.g. situation awareness or skill degradation) in NPP operations \cite{jo2012quantitative}.

\begin{table}[h]
\caption{Comparison of Mechanistic Simulation: IDAC vs. QN-MHP vs. ACT-R}\label{mechanistic}%

\begin{tabular}{p{0.15 \linewidth} p{0.25\linewidth} p{0.28\linewidth} p{0.28\linewidth}}

\toprule
\textbf{Metrix}  & \textbf{IDAC} & \textbf{QN-MHP}  & \textbf{ACT-R}\\
\midrule
\textbf{Modeling scope and fidelity} & 
 IDAC offers a higher-level yet holistic crew-centric model embedding cognitive, emotional, and team dynamics specific to nuclear operational contexts.
& 
QN-MHP occupies a middle ground, focusing on the temporal concurrency of tasks and workload rather than internal knowledge states, which makes it well-suited for evaluating interface designs and staffing strategies but less so for simulating complex diagnostic reasoning. &  provides the most granular simulation of internal cognitive processes (memory chunks, problem-solving steps, attention shifts), aligning closely with human data\\[1ex]
\midrule
\textbf{Practical application and validation} & 
IDAC has seen success in risk assessment case studies, directly feeding into human reliability predictions in PRA with its probabilistic outputs.
& 
QN-MHP has been validated in experimental settings for predicting performance metrics (e.g. task completion time, error rates) and thus serves as a decision-support tool for HFE improvements. & 
 ACT-R, supported by a large research community, boasts a wide array of validations across domains; in nuclear scenarios, it has demonstrated predictive accuracy for both operator performance and underlying mental workload (e.g. via physiological correlates).\\[1ex]
\midrule
\textbf{Interpretability and explainability} & 
IDAC’s rule-based structure offers transparent logic for crew actions (mirroring procedural steps and known error mechanisms), but it may abstract away individual cognitive variability. 
& 
QN-MHP outputs are easily interpreted in terms of delays and queue lengths (i.e. where bottlenecks in human processing occur), providing clear performance insights, though they might not explain why a bottleneck arises cognitively.  & ACT-R models, while more complex, can explain behavior in terms of cognitive theories (e.g. memory decay causing an omission) and even predict brain activation patterns, lending a form of explanatory depth and consistency with neuroscience.\\
\midrule

\textbf{Computational demands} & computationally intensive&  computed relatively faster &  computed relatively faster \\
\midrule
\textbf{Integrability and extensibility} & IDAC has evolved mainly within the nuclear domain and is tightly integrated with specific simulation platforms (ADS), making it highly specialized but less flexible for other uses.& QN-MHP, grounded in general HCI modeling, can be adapted to various multitasking scenarios, though it lacks a built-in learning mechanism to extend into long-term adaptations of operator behavior. & ACT-R stands out for its modular integration – researchers have linked it with other models (vision systems, robotic interfaces, etc.) and extended it (e.g. ACT-R$\Phi$ for physiology) to broaden its applicability.  \\

\botrule
\end{tabular}
\end{table}

Each architecture brings a unique perspective to modeling nuclear operator behavior. The characteristics of each method can be found in the table \ref{mechanistic}. In summary, the past decade has seen substantial progress in mechanistic simulation of nuclear power plant operator behavior using cognitive architectures. IDAC has matured as a domain-focused tool capable of simulating crew decisions and errors in high-fidelity accident scenarios, providing valuable inputs to probabilistic risk assessments. QN-MHP has emerged as an effective approach to quantifying human performance and mental workload in complex multitasking settings, aiding in HMI design and staffing evaluations by predicting where cognitive overload may occur. ACT-R, leveraging its deep roots in cognitive science, offers unparalleled cognitive accuracy and theoretical grounding, now demonstrated in nuclear-related use cases that demand detailed modeling of human reasoning and behavior under stress. Notably, ACT-R’s alignment with human cognition (both behaviorally and neurologically) and its extensibility make it a compelling choice for integrated human-in-the-loop simulations, despite the effort required to harness it. Therefore, ACT-R was used for cognitive modeling in this study.

\section{Methodology}\label{Methodology}

\subsection{The Proposed Framework COGMIF}

\begin{figure}[h]
\centering
\includegraphics[width=1.0\textwidth]{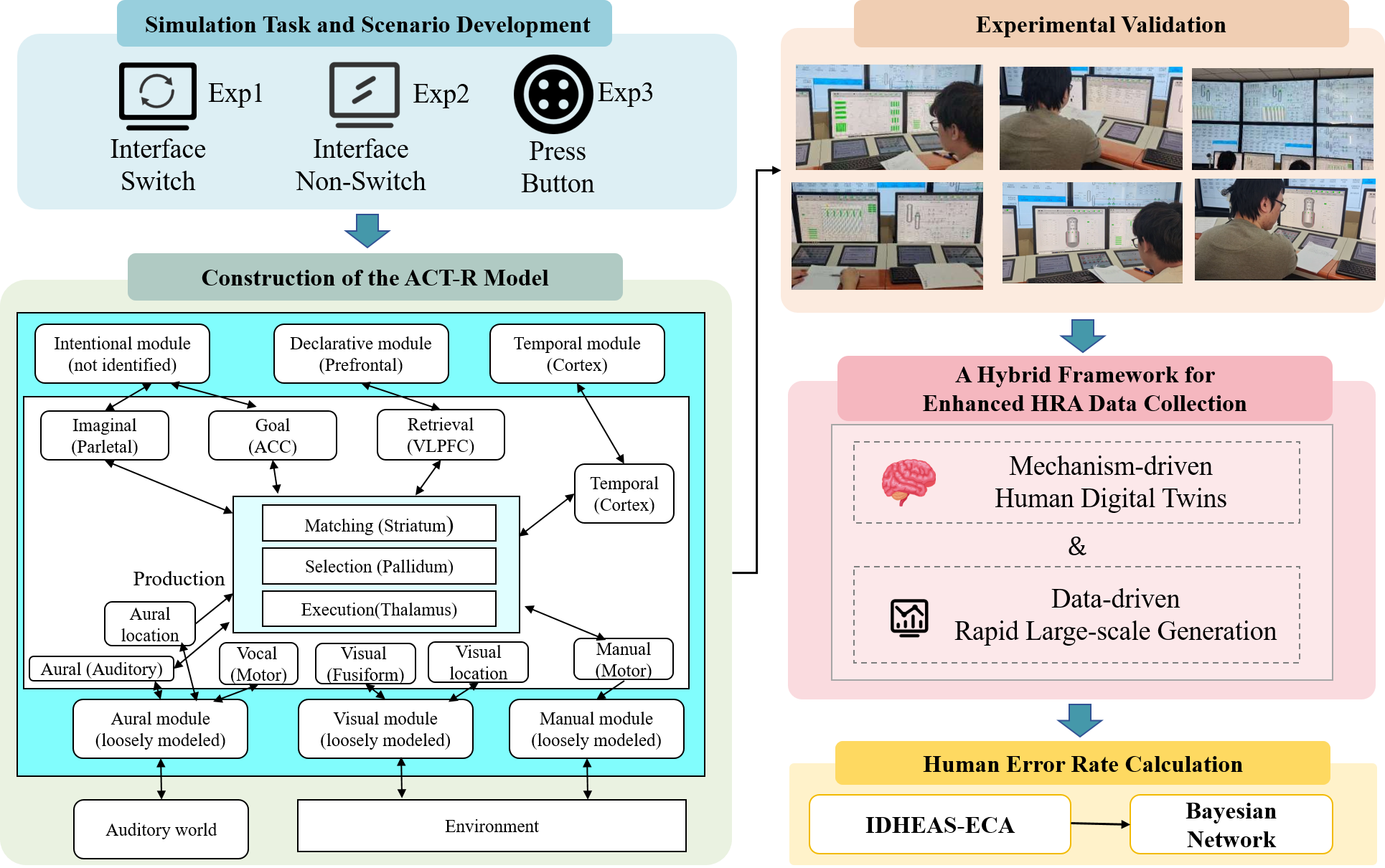}
\caption{The Proposed Framework COGMIF}\label{workflow}
\end{figure}

The methodology framework of this study, as illustrated in Figure~\ref{workflow}, comprises four interconnected phases: (1) simulation task and scenario development, (2) ACT-R cognitive architecture modeling, (3) experimental validation, (4) a hybrid framework for enhanced HRA data collection, and (5) human error rate calculation.
This structure systematically investigates human reliability factors in safety-critical systems through computational modeling and empirical verification. The initial phase establishes the operational context via rigorous scenario design, using selected abnormal operating procedures as case studies. The second phase employs the ACT-R architecture to construct mechanistically driven human digital twins, modeling cognitive processes under complex and high-stress conditions. The third phase validates model predictions by collecting human performance metrics and performing comparative analyses against simulated outcomes. In the fourth phase, a hybrid simulation framework is introduced to address the limitations of manual modeling and to improve the scalability of HRA data generation. This hybrid integration of mechanistic (ACT-R) and data-driven (TimeGAN) approaches allows for both interpretability and scalability, offering a practical and efficient alternative to traditional human-in-the-loop experiments. In the final stage, the data generated from the phase IV hybrid framework are utilized to compute the time-based failure probability ($P_{t}$) within the IDHEAS-ECA methodology. Additionally, these outputs are mapped onto a Bayesian network to facilitate the estimation of human error rates through probabilistic reasoning.

Technical specifications and implementation details for each component are elaborated in Sections \ref{Stage 1}–\ref{Stage 5}.

\subsection{Stage 1: Simulation Task and Scenario Development} \label{Stage 1}

In this section, some straightforward operational processes were selected for experimentation. Given the unique demands of nuclear power plant operations, every operator action must be accompanied by a three-way communication with the shift supervisor. To simulate this condition, our volunteer engaged in such triadic interactions. Consequently, errors were seldom observed in this experimental setup, and our focus was confined to evaluating the temporal plausibility of the ACT-R simulation. Additionally, two supplementary experiments were designed that required instantaneous responses from the operators, thereby enabling the collection of error data.

\begin{figure}[h]
\centering
\includegraphics[width=1.0\textwidth]{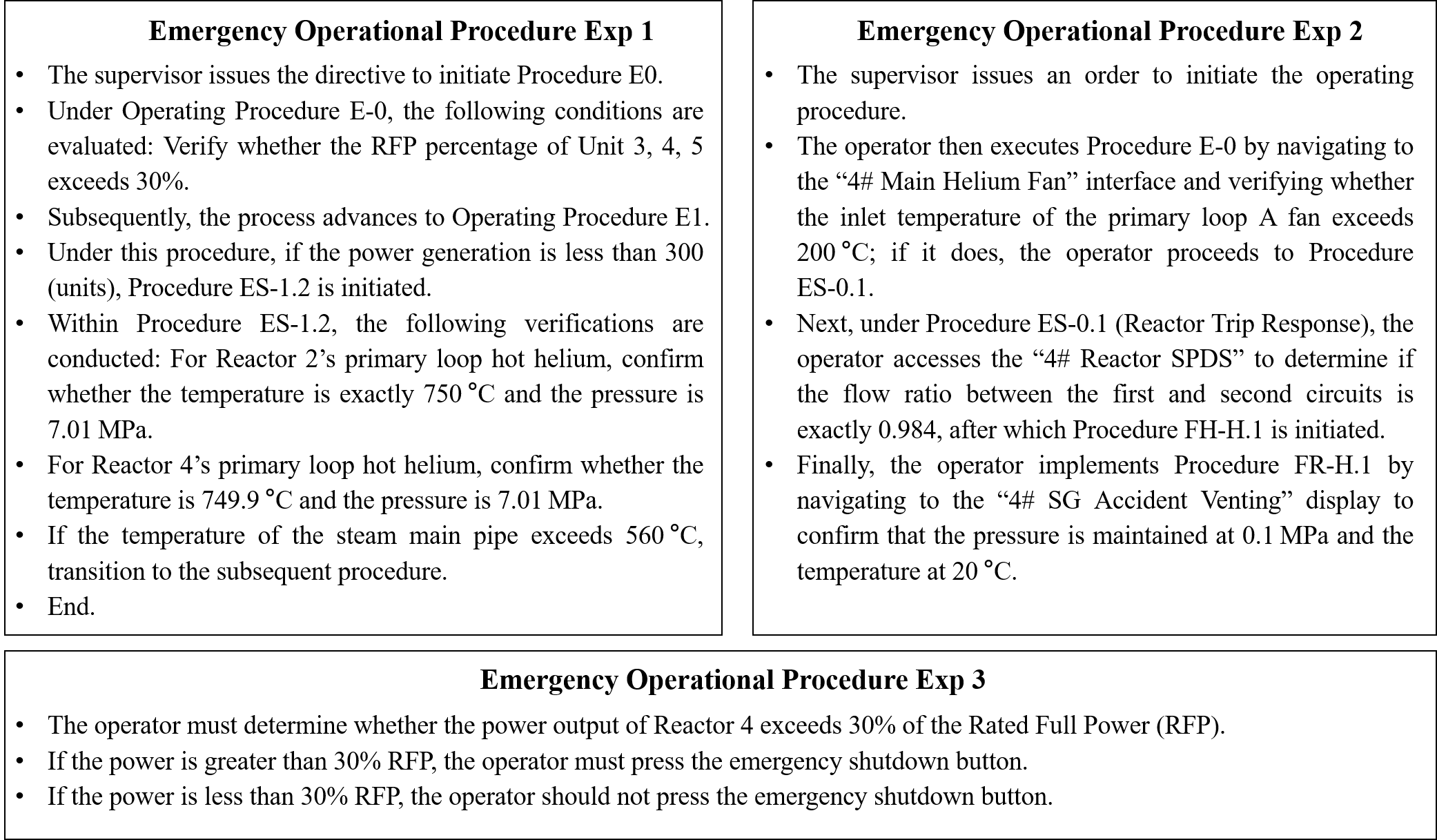}
\caption{Content of Experimented Emergency Operational Procedures}\label{EOPs}
\end{figure}

As for time estimation, the experiment involves implementing response procedures under abnormal conditions, as depicted in Figure \ref{EOPs}. Three experimental conditions were established: Experiment 1 entails interface switching; Experiment 2 does not require interface switching; and Experiment 3 involves a manual action (i.e., pressing a button).

\subsection{Stage 2: Construction of the ACT-R model}

ACT-R is a comprehensive cognitive architecture developed to model human cognition and simulate the underlying processes involved in various cognitive tasks \cite{lebiere2019error}. It is one of the most prevailing cognitive architectures based on a rigorous theory of human cognition during the last three decades. It is based on the neuroscience theory. The functions of modular organization of ACT-R can be mapped onto some brain regions \cite{taatgen2006modeling}. The ACT-R cognitive architecture fundamentally distinguishes between declarative and procedural knowledge, which accords with the literature on skill acquisition in philosophy and psychology. It is based on the "if-then" rule and has been applied in various fields, such as computational cognitive modeling of pilot performance in preflight and takeoff procedures \cite{cao2024computational}, quantitative prediction of mental workload \cite{jo2012quantitative}, and interface design \cite{byrne2001act}.

ACT-R operates on the premise that human cognition can be understood as the interaction between a declarative memory system \cite{eichenbaum2000cortical} (containing facts and knowledge) and a procedural memory system \cite{tulving1985many} (containing cognitive skills and strategies), with both systems working together to produce goal-directed behavior. The architecture is particularly well-suited for simulating complex cognitive processes and has been widely used in human factors research \cite{atashfeshan2017determination}, cognitive modeling \cite{jo2012quantitative}, and human-computer interaction studies \cite{cao2024computational}.

\begin{figure}[h]
\centering
\includegraphics[width=1.0\textwidth]{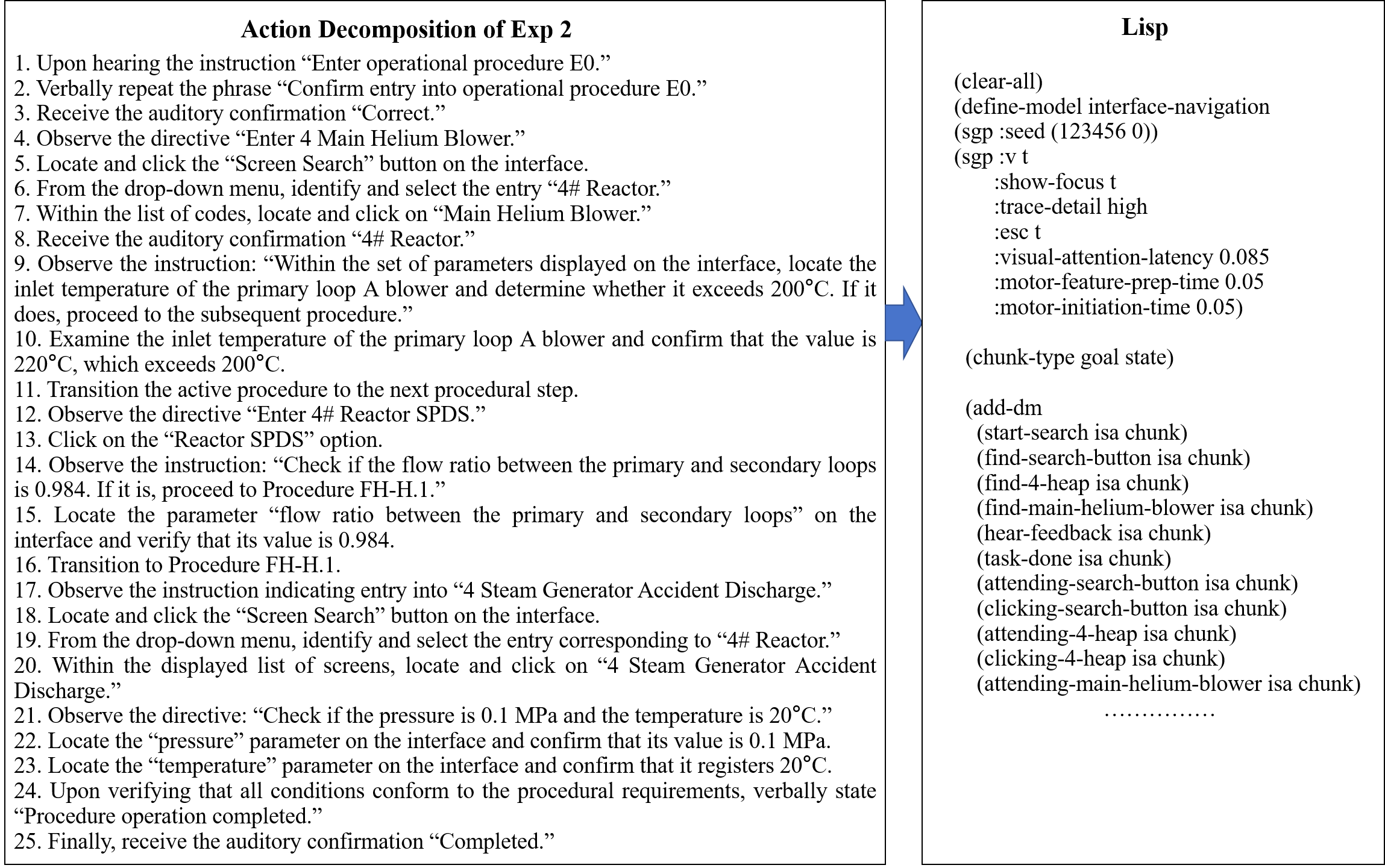}
\caption{Integrated Schematic Diagram of Exp2 Basic Operations and Refined Protocols for LISP Model Construction}\label{lisp}
\end{figure}

The ACT-R cognitive architecture model was implemented using the Lisp programming language. This required decomposing complex operational tasks into granular procedural branches. The most intricate experimental task (Exp2) was selected as a representative case for detailed modeling. As illustrated in Figure \ref{lisp}, the left panel demonstrates the refined operational protocol of Exp2, while the right panel presents the corresponding Lisp code implementation. This dual representation systematically maps procedural requirements to executable cognitive operations within the ACT-R framework.

\begin{figure}[h]
\centering
\includegraphics[width=1.0\textwidth]{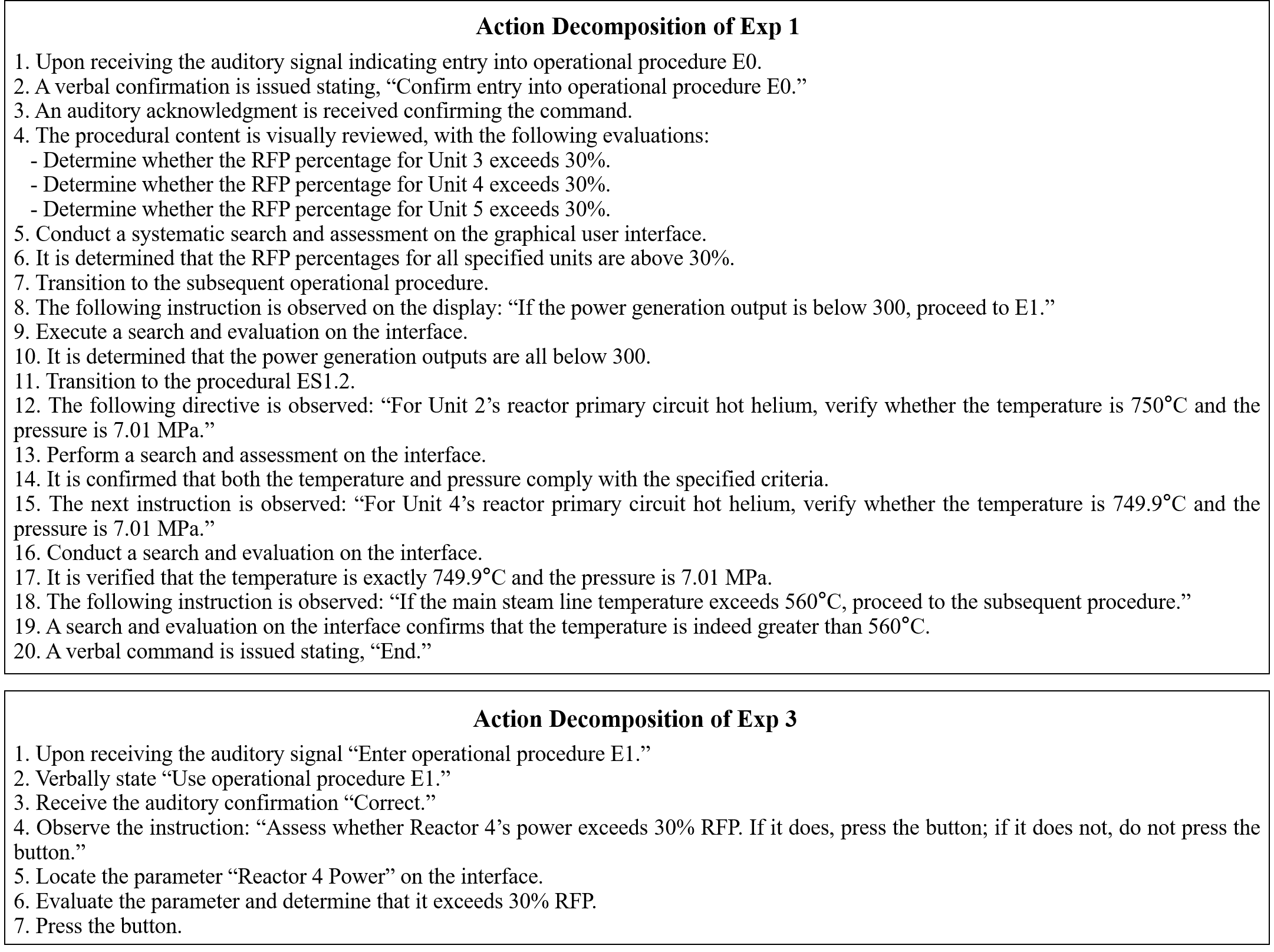}
\caption{Refined Protocols of Exp1 and Exp3 for LISP Model Construction }\label{decomposition}
\end{figure}

For simpler experimental tasks (Exp1 and Exp3), analogous decomposition was performed as shown in Figure \ref{decomposition}. The modeling process systematically addressed six core cognitive components: (1) auditory processing of system alerts, (2) verbal response generation, (3) button press operations, (4) decision-making mechanisms, (5) parameter retrieval from interface elements, and (6) procedural switching logic. Each cognitive operation was encoded as production rules with associated latency parameters derived from empirical human performance data.

All model implementations incorporated ACT-R's default temporal parameters for basic cognitive operations. Task-specific adjustments were made through declarative memory chunk activation levels and procedural rule conflict resolution mechanisms. This multi-layered approach enabled the model to simulate both routine operations and exception-handling scenarios observed in human operators.

\subsection{Stage 3: Experimental Validation on HTGR}

\begin{figure}[h]
\centering
\includegraphics[width=0.6\textwidth]{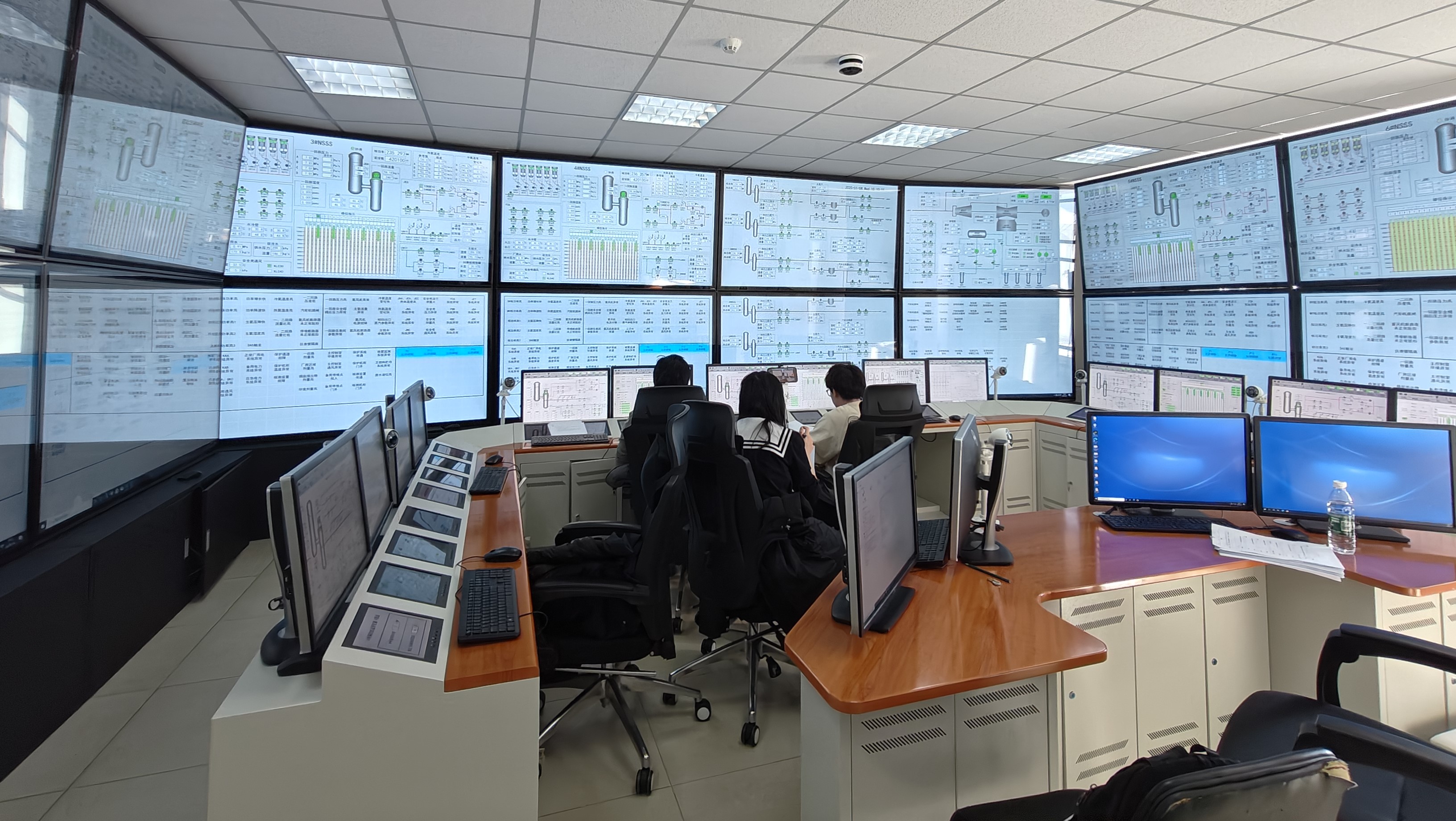}
\caption{HTGR Simulated Main Control Room at Tsinghua University's Institute of Nuclear Energy Technology (1:1)}\label{simulator}
\end{figure}

The high-temperature gas-cooled reactor (HTGR) is a multi-module reactor design, which has already been implemented in Shidao Bay, Shandong, China. This reactor module consists of six reactors: reactor 1-6. In the main control room, there are five operators: three reactor operators (one for each pair of reactors), one mechanical operator, and one shift supervisor. Each reactor operator is responsible for two reactors. We gathered three graduate students to conduct experiments on a 1:1 scale simulator for HTGR shown in Figure \ref{simulator}. First, we developed comprehensive experimental protocols in advance and implemented a structured training program focusing on interface familiarization. Subsequently, we conducted experimental sessions in controlled individual settings, with all operator interactions documented through synchronized video recording systems. For data collection, three distinct protocols were implemented, Exp1-Exp3, with sample sizes of 5, 5, and 12, respectively. 

\subsection{Stage 4: A Hybrid Framework for Enhanced HRA Data Collection}

The ACT-R model provides a mechanism-based simulation of operator cognitive processes, including perception, memory retrieval, decision-making, and motor execution. While this ensures interpretability and fidelity to human cognitive mechanisms, generating sufficient data for large-scale analysis remains a bottleneck due to the computational cost of simulating individual task episodes. 

To overcome this, this study proposes a hybrid framework that integrates mechanism-driven knowledge with data-driven methods. First, we utilize the ACT-R model to simulate operator behavior and generate time-related data samples, which are relatively easier to predict and model. These sampled time data are then used to train a generative neural network based on TimeGAN. This approach enables the synthesis of large-scale, realistic time-series data from a limited number of experimental simulations.

\begin{figure}[h]
\centering
\includegraphics[width=0.6\textwidth]{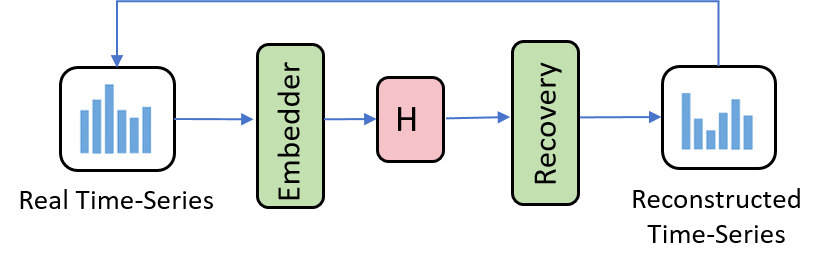}
\caption{The Part 1 of the TimeGAN Model for Synthetic Time-Series Generation}\label{stage1}
\end{figure}

The part 1 of TimeGAN is illustrated in Figure~\ref{stage1}. In this stage, the model aims to learn a meaningful latent representation of the real-time-series data. Specifically, the embedder module maps the real time-series data into a latent space representation H. This module is implemented using a recurrent neural network (RNN) architecture, and in this study, we adopt the gated recurrent unit (GRU) due to its effectiveness in capturing temporal dependencies. The recovery module then reconstructs the time-series data from the latent representation H, resulting in the reconstructed time-series. This stage focuses on training the embedder and recovery components jointly to ensure that the latent space preserves the essential temporal structure of the original data.

\begin{figure}[h]
\centering
\includegraphics[width=0.6\textwidth]{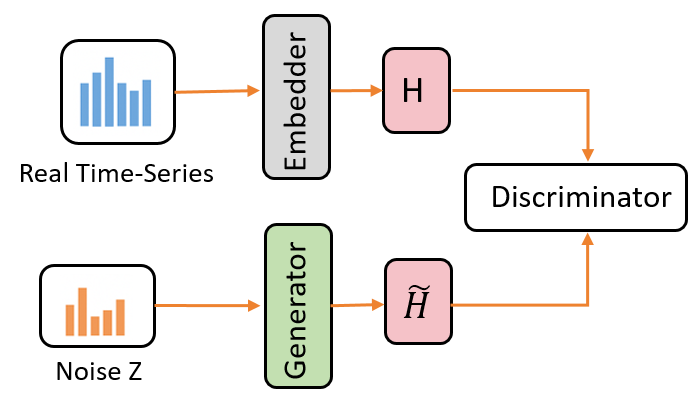}
\caption{The Part 2 Architectural of the TimeGAN Model for Synthetic Time-Series Generation}\label{stage2}
\end{figure}

The part 2 of TimeGAN is illustrated in Figure~\ref{stage2}. In this stage, the training primarily involves the generator and the discriminator modules in an adversarial manner, while the embedder trained in Stage 1 is kept fixed. A random noise tensor Z is fed into the Generator, which learns to produce fake latent representations $\tilde{H}$ that mimic the distribution of the real latent representations H. The Discriminator is trained to distinguish between the real latent representations H, obtained from the fixed embedder, and the generated ones $\tilde{H}$. This adversarial training encourages the Generator to produce increasingly realistic latent representations, thereby enhancing the quality of the synthesized time-series data.

This hybrid framework offers the following benefits: (i) Enables generation of large volumes of synthetic data for diverse operating scenarios, (ii) Retains the cognitive realism established by the ACT-R-based digital twin, (iii) Reduces the computational burden associated with repeated mechanistic simulations. The integration of TimeGAN thus complements the mechanistic modeling of ACT-R, forming a synergistic simulation environment that advances the scope and depth of HRA studies. 

\subsection{Stage 5: Human Error Rate Calculation}\label{Stage 5}

\begin{figure}[h]
\centering
\includegraphics[width=0.8\textwidth]{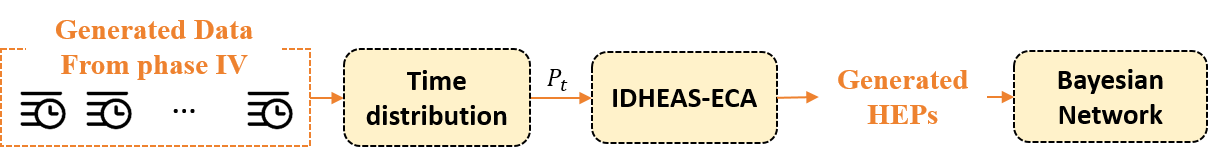}
\caption{ The Human Error Rate Calculation Module}\label{hEP}
\end{figure}

The human error rate calculation module in COGMIF is shown in Figure \ref{hEP}. In this stage, the time-series data generated by the hybrid framework are fed into the IDHEAS-ECA framework as the distribution of required time to derive corresponding HEPs. Finally, to validate the plausibility and robustness of the generated HEPs, we perform a comprehensive analysis using a Bayesian network. This includes influence strength evaluation, and sensitivity analysis, ensuring the reliability and applicability of the synthesized error data in real-world HRA contexts.

\section{Results and Evaluation}\label{Results and Evaluation}

\subsection{Temporal Validation and Comparative Analysis}

In order to assess the temporal validity of the HDT constructed using ACT-R, we simulated the entirety of Exp1-Exp3 and identified several key time points within these experiments. The corresponding simulation times were then extracted to facilitate a comparison between the experimental results and the simulation outcomes.

\begin{figure}[h]
\centering
\includegraphics[width=0.8\textwidth]{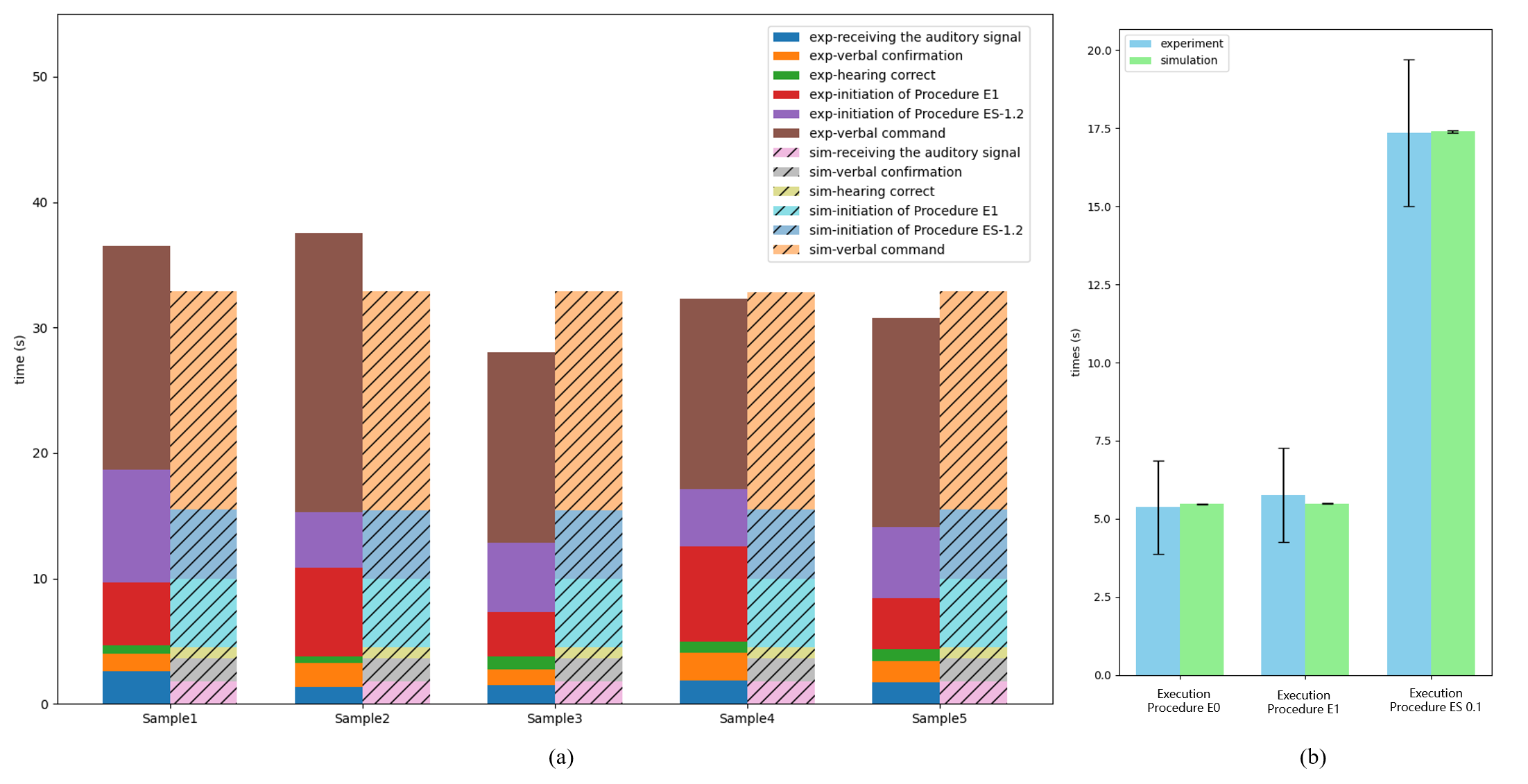}
\caption{Temporal Analysis of Experimental and Simulated Procedures in Exp1}\label{exp1}
\end{figure}

Figure \ref{exp1} (a) illustrates the five experimental trials of Exp1, comparing the duration of each procedural step as well as the cumulative time between the simulation and the actual experiment. The comparison indicates that both the overall and the individual component times are in close agreement. In the right panel (b), the figure presents a comparison of the mean and variance for two specific segments: (i) execution Procedure E-0,  (ii) execution Procedure E-1, and (iii) execution Procedure ES-1.2. For the first segment, the actual mean duration is 5.3722 s while the simulated mean is 5.4777 s, with variances of 1.4922 and 0.0139, respectively. For the second segment, the actual and simulated means are 5.7667 s and 5.485 s, respectively, with corresponding variances of 1.5103 and 0.01136. For the last segment, the actual and simulated means are 17.35 s and 17.39 s, respectively, with corresponding variances of 2.338 and 0.0423. These results suggest that, overall, the mean durations are largely consistent between the experimental data and the ACT-R simulation, although the simulation exhibits significantly lower variance.

\begin{figure}[h]
\centering
\includegraphics[width=0.8\textwidth]{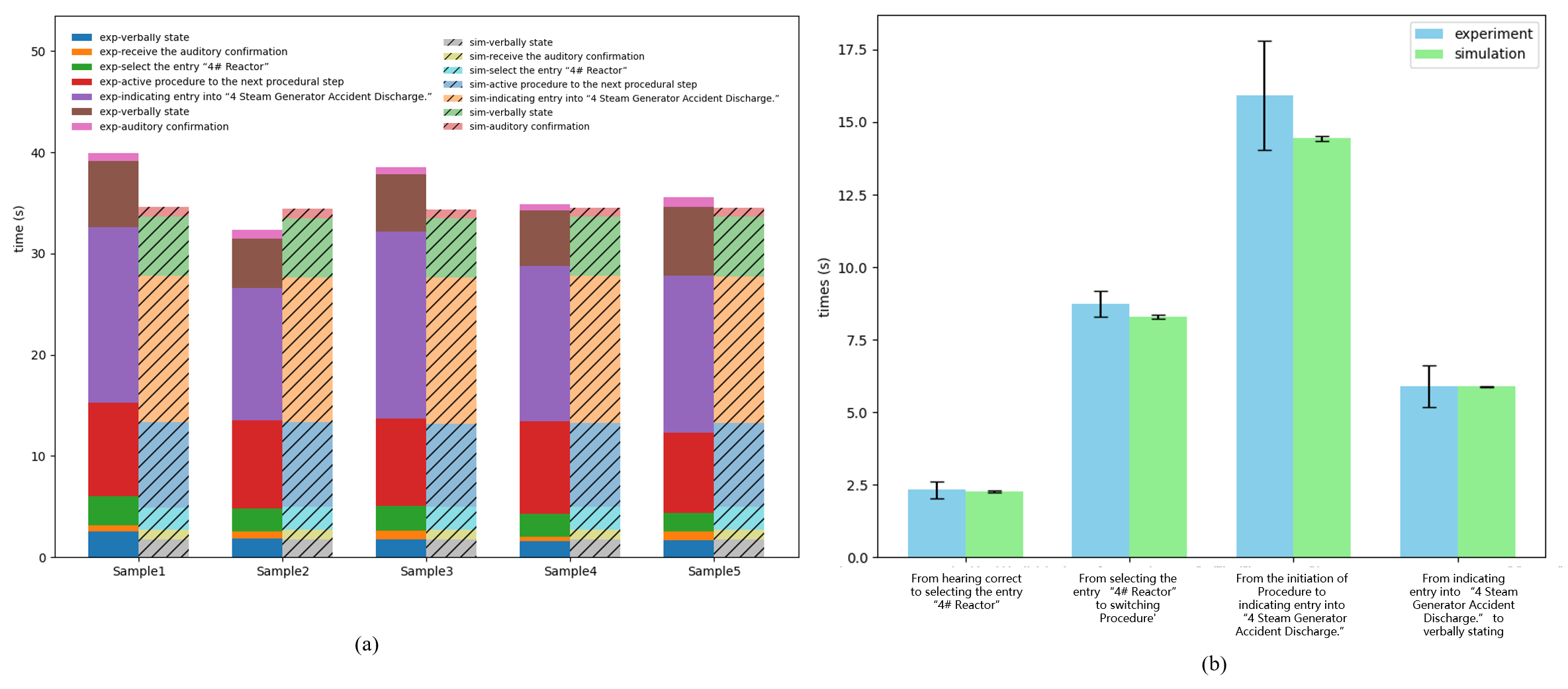}
\caption{Temporal Analysis of Experimental and Simulated Procedures in Exp2}\label{exp2_compare}
\end{figure}

Figure \ref{exp2_compare} (a) presents the results from five experimental trials of Exp2, comparing the durations of individual procedural steps as well as the cumulative time between the simulation and the actual experiment. The comparison demonstrates that both the overall total time and the time allocated to each step are consistent. Figure \ref{exp2_compare} (b) provides a comparative analysis of the means and variances for four specific processes: (i) from hearing the correct stimulus to selecting the entry “4\# Reactor,”, (ii) from selecting the entry “4\# Reactor” to switching procedure, (iii) from the initiation of Procedure to indicating entry into “4 Steam Generator Accident Discharge,” and (iv) from indicating entry into “4 Steam Generator Accident Discharge” to verbally stating. For these processes, the actual mean durations are 2.3200 s, 8.7333 s, 15.9200 s, and 5.9000 s, respectively, while the corresponding simulated means are 2.2656 s, 8.2916 s, 14.4428 s, and 5.8858 s. The actual variances are 0.3030, 0.4551, 1.8847, and 0.7121, compared to the simulated variances of 0.0300, 0.0668, 0.0833, and 0.0098. Overall, the mean durations are largely consistent between the simulation and the experimental results; however, the ACT-R simulation exhibits considerably lower variance.

\begin{figure}[h]
\centering
\includegraphics[width=0.8\textwidth]{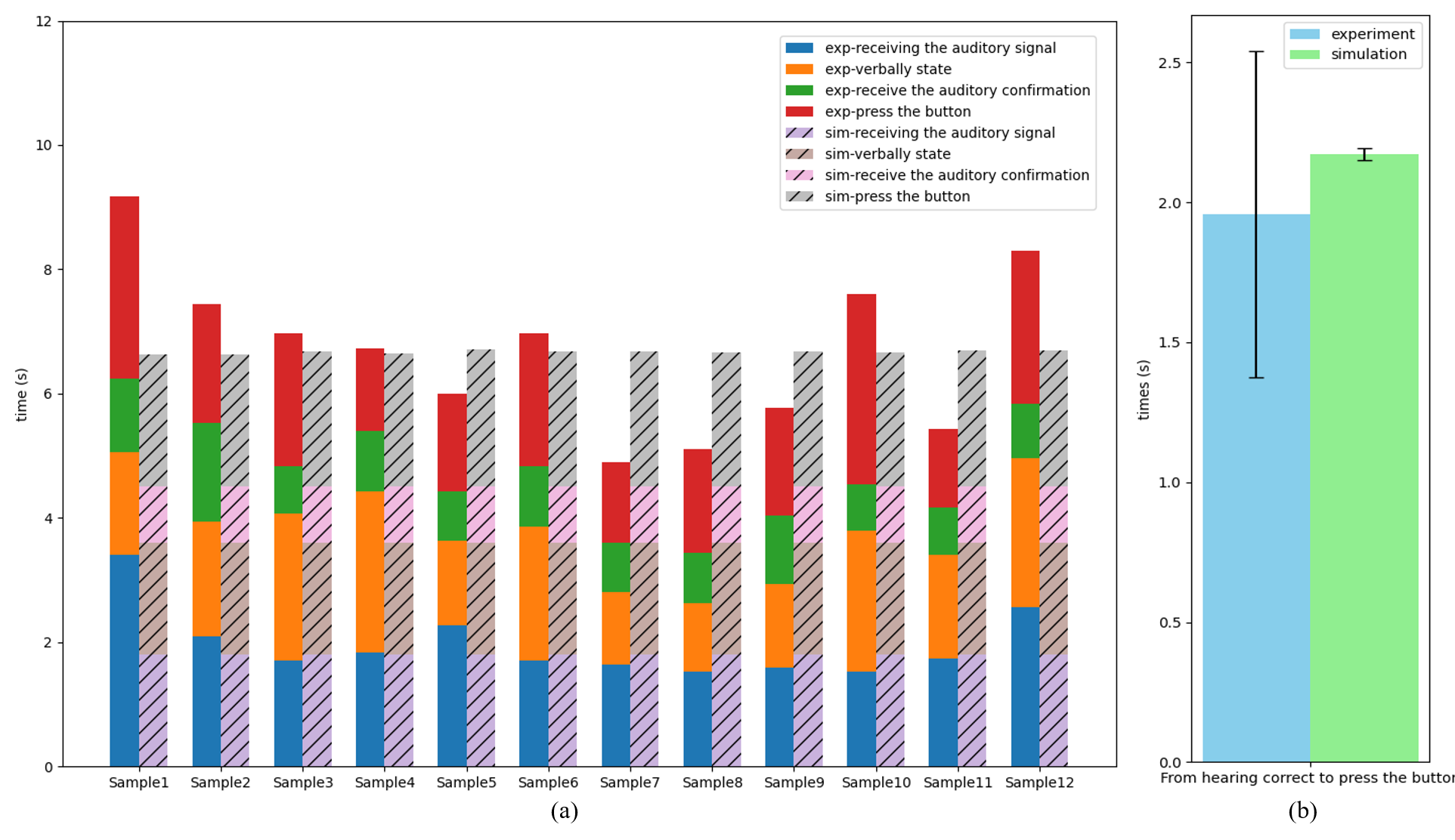}
\caption{Temporal Analysis of Experimental and Simulated Procedures in Exp3}\label{exp3}
\end{figure}

Figure \ref{exp3} (a) illustrates the 12 experimental trials of Exp3, comparing the durations of individual procedural steps and the cumulative time between the simulation and the actual experiment. The results indicate that both the overall total time and the durations of the individual segments are consistent. In the right subfigure (b), a comparison of the mean and variance for the process "From hearing correctly to pressing the button" is presented. The actual mean is 1.9583 s, while the simulated mean is 2.1723 s, with corresponding variances of 0.5841 and 0.0232, respectively. Overall, the mean values are closely aligned, though the ACT-R simulation exhibits a notably lower variance.

\subsection{ Enhancing Data Generation Using TimeGAN}\label{Enhancing Data Generation Using TimeGAN}

One current limitation of our method lies in the manual nature of the simulation process, which can be time-consuming and labor-intensive. To address this, we aim to leverage AI-based algorithms to enable large-scale data generation. Taking Exp1 as an example, we employed the TimeGAN algorithm, a two-stage training framework. This hybrid architecture not only generates time series data that appear realistic, but also preserves the intrinsic temporal dependencies.

In the following analysis, we focus on three segments from Exp 1: "Execution Procedure E-0" (S1), "Execution Procedure E-1" (S2) and "Execution Procedure ES-1.2" (S3). These segments are used to train a TimeGAN model. We supplemented the dataset with 40 simulated trials generated using the ACT-R cognitive architecture. After training the TimeGAN model with this expanded dataset, we conducted testing and presented below ten representative sequences from the model's output.

\begin{figure}[h]
\centering
\includegraphics[width=0.8\textwidth]{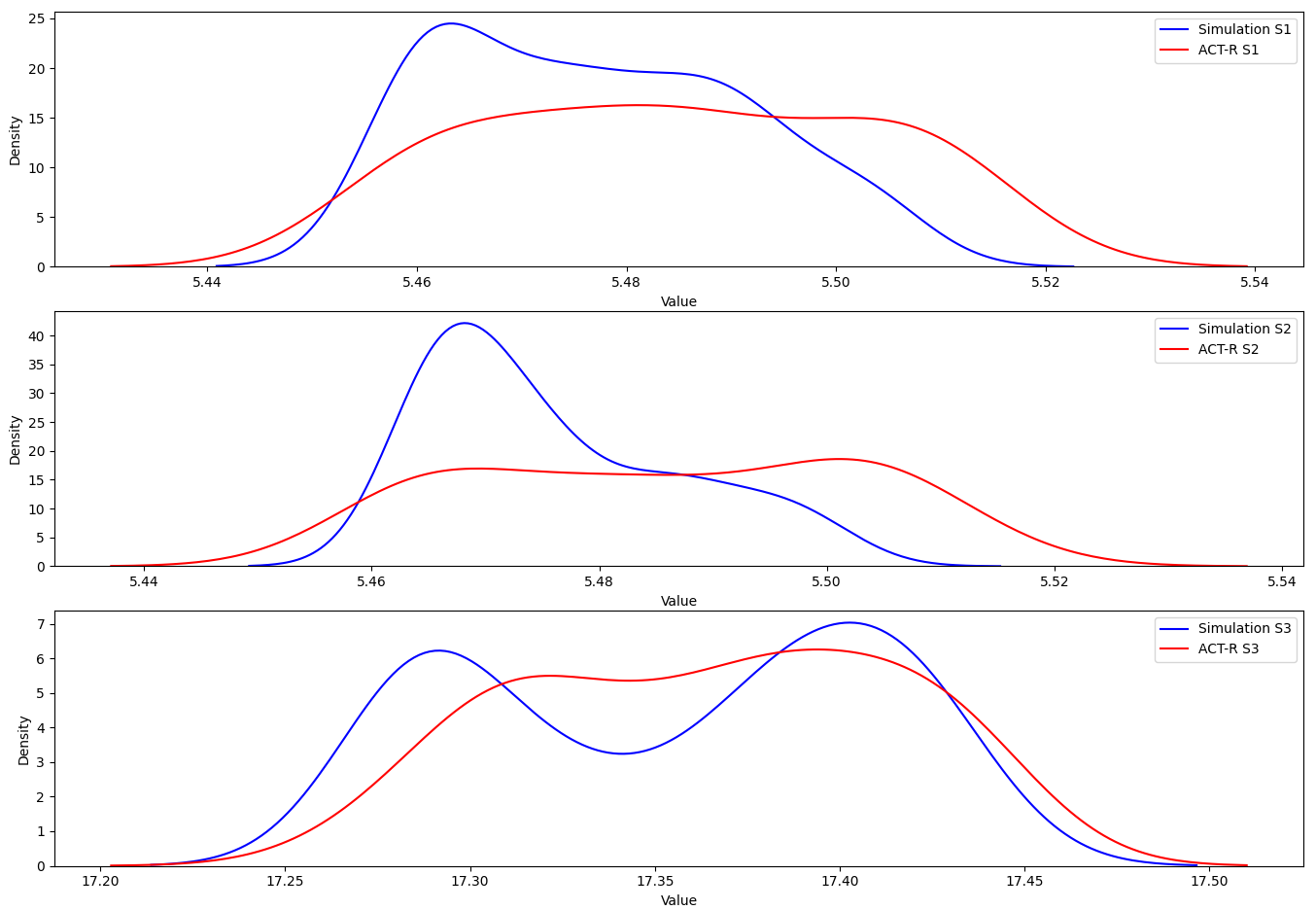}
\caption{Comparison of Response Time Distributions using Kernel Density Estimation (KDE)}\label{compare_kde}
\end{figure}

As shown in Figure \ref{compare_kde}, the result shows the kernel density estimation (KDE) plots comparing the TimeGAN-generated simulation data (blue curves) with the ACT-R model outputs (red curves) for task duration across S1,to S3 in Exp1. In S1, both distributions exhibit unimodal structures with comparable ranges, although the simulated distribution is more peaked and slightly shifted to the left relative to the ACT-R output. This suggests that TimeGAN captures the central tendency well. In S2, a similar pattern is observed. As for S3, the distributions show greater divergence, with both displaying a bimodal structure. The synthetic simulation captures the general shape of the ACT-R output but differs in the relative height and position of the two peaks. 

In addition to the KDE graphical visualization, several quantitative metrics were employed to evaluate the consistency between simulation and experimental data. These metrics include the mean squared error (MSE), mean absolute error (MAE), and coefficient of variation (CV).

The MSE, as defined in Equation \ref{MSE}, quantifies the average of the squared differences between the simulated values $(y_{sim}^{i }$ and the corresponding experimental values $ y_{exp}^{i }$, where n denotes the total number of data points:
\begin{equation}\label{MSE}
    MSE=\frac{1}{n}\sum_{i=1}^{n}(y_{sim}^{i } - y_{exp}^{i } )  ^{2}
\end{equation}

Similarly, the MAE measures the average magnitude of the absolute errors between the simulation and experimental results, offering an interpretable scale of deviation. The MAE is calculated as shown in Equation \ref{MAE}:

\begin{equation} \label{MAE}
    MAE=\frac{1}{n} \sum_{i=1}^{n}\left | y_{sim} ^{i} - y_{exp}^{i}  \right |  
\end{equation}

The CV is a metric used to assess the relative dispersion of data by comparing the standard deviation to the mean. It provides an understanding of the degree of variability in relation to the size of the mean. The CV is calculated as:

\begin{equation}\label{CV}
    CV=\frac{\sigma }{\mu } 
\end{equation} 

where $\sigma$ represents the standard deviation of the data, and $\mu$ denotes the mean of the data. This ratio is particularly useful when comparing the variability of datasets with different units or scales.

\begin{table}[h]
\caption{Quantitative Comparison Between TimeGAN-Generated and ACT-R-Based Data Using MAE, MSE, and CV Metrics}\label{Metrix}%
\begin{tabular}{p{0.15\linewidth} p{0.15\linewidth} p{0.15\linewidth} p{0.15\linewidth}}
\toprule
\textbf{Step} & \textbf{MAE} & \textbf{MSE}  & \textbf{CV} \\
\midrule
S1 & 0.000449&  0.021195 & 0.001311 \\
S2 & 0.000314&  0.017716 & 0.001400 \\
S3 &  0.004490& 0.067008& 0.000787 \\
\bottomrule
\end{tabular}
\end{table}

The results in Table~\ref{Metrix} demonstrate that for S1 to S3, the TimeGAN model achieves very low error values, with MAE below 0.0005 and MSE under 0.022, indicating a high level of fidelity in replicating the underlying temporal patterns. The CV values, which reflect relative variability, are also consistently low, on the order of $1e-3$, confirming the model’s ability to preserve the dispersion characteristics of the original data. For S3, both MAE and MSE are slightly higher (0.00449 and 0.067, respectively), suggesting a marginal decrease in generation accuracy, which may be attributed to greater variability or complexity in the original task duration distribution for that step. Nonetheless, the CV remains below 0.001, indicating that the relative structure of the data is still well maintained.

Overall, the TimeGAN model demonstrates strong predictive performance in capturing the response time distributions originally generated by the ACT-R cognitive model. The KDE comparisons indicate good alignment in terms of both distributional shape and central location, validating the model’s utility for simulating human-like temporal behavior in procedural task environments.

\subsection{ Human Error Probability Generation Using IDHEAS}%HEP的对比

The human reliability assessment model evaluates human errors in cognitive tasks via two principal mechanisms, errors of omission (EOO) and errors of commission (EOC) \cite{xiao2025dynamic}. As for EOO, each memory chunk is assigned an activation value that governs both the speed and likelihood of its retrieval. ACT-R posits the existence of a retrieval delay threshold; if a memory chunk cannot be retrieved within the prescribed time window due to insufficient activation, compounded by the addition of Gaussian noise, the retrieval process fails, resulting in an omission error. This mechanism mirrors the tendency for information to be overlooked when the memory load is high or when the task complexity is elevated.

As for EOC, in addition to complete retrieval failures, the ACT-R model permits “partial matching.” Under this mechanism, even if a memory chunk does not fully align with the expected criteria, it may still be erroneously retrieved provided that its activation value remains sufficiently high (after the application of penalties for mismatches), leading to a commission error. For example, in arithmetic tasks, this mechanism accounts for the phenomenon whereby common errors are numerically close to the correct answer.

To verify and demonstrate the accuracy of the generated error probabilities, we selected Exp1 as a representative case study. The specific procedural flow of Exp1 is illustrated in Figure~\ref{EOPs}. The core of this operational sequence involves the execution of three critical procedures: E-0, ES-1, and ES-1.2.

For the error analysis, we adopt the state-of-the-art IDHEAS-ECA method. Developed by the U.S. Nuclear Regulatory Commission (NRC), IDHEAS-ECA is currently one of the most authoritative and systematically validated methods for HRA, supported by robust data foundations and well-documented case studies. In the IDHEAS-ECA framework, the overall human error probability $P_{HFE}$ is calculated by jointly considering cognitive failure modes and time constraints. The computation is formalized as shown in Equation~\ref{IDHEAS}:

\begin{equation} \label{IDHEAS}
    P_{Event} = 1-(1-P_{c}) \cdot (1-P_{t})
\end{equation}

Where, $P_{c}$ represents the sum of human error probabilities associated with cognitive failure modes, capturing the likelihood of failures due to issues such as detection, understanding, decisionmaking, action execution, and interteam coordination.  On the other hand, $P_{t}$ denotes the convolution of the distribution of time available and time required, reflecting the probability of failure due to insufficient time to complete a task. This term effectively captures the impact of temporal pressure on operator performance.

For the computation of $P_{c}$, we conducted a detailed analysis based on the publicly available guidelines provided in the IDHEAS-ECA methodology. According to the principles and documentation of the IDHEAS-ECA methodology, we defined specific human failure events (HFEs) for Exp1. The details of these HFEs are presented in Table~\ref{HFEs}.

\begin{table}[h]
\caption{Identified Human Failure Events (HFEs) for Key Procedures in Exp1}\label{HFEs}%
\begin{tabular}{p{0.1\linewidth}  p{0.7\linewidth} }
\toprule
\textbf{Procedure} & \textbf{HFE}  \\
\midrule
E-0 & Operators fail to verify the RFP\% values of pumps 3, 4, and 5\%, or omit transitioning to E-1 despite all conditions being met. \\
E-1 & Operators fail to monitor and respond to power level changes, leading to the omission of ES-1.2 entry when power drops below 300 MW. \\
ES-1.2 & Operators fail to assess or act upon deviations in helium loop temperature/pressure and steam header temperature, missing required transitions. \\
\bottomrule
\end{tabular}
\end{table}

\begin{figure}[h]
\centering
\includegraphics[width=0.9 \textwidth]{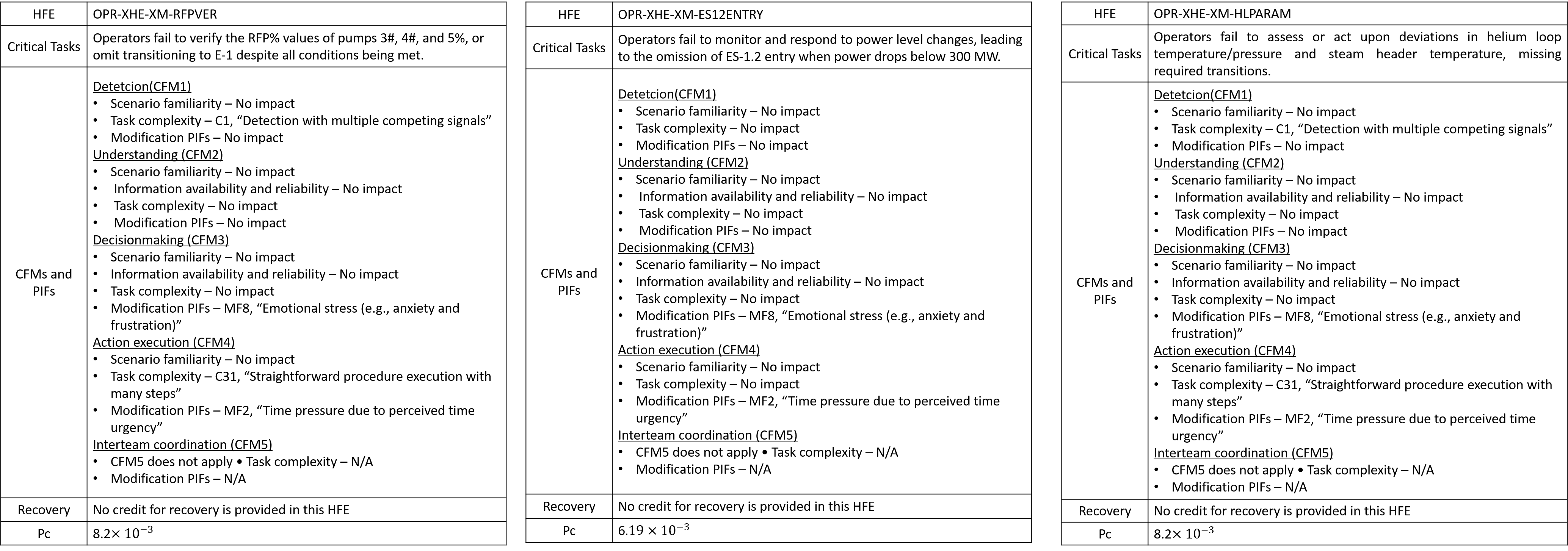}
\caption{IDHEAS-ECA Analysis Worksheets for S1 to S3}\label{IDHEAS_sheet}
\end{figure}

The results of the $P_{c}$ estimation, derived through this approach, are summarized in Table~\ref{Ps}. The detailed step-by-step analysis used to calculate $P_{c}$ is shown in Figure~\ref{IDHEAS_sheet}. In addition, we conducted a comparative analysis using the current industry-standard SPAR-H method. The corresponding HEP estimates for Steps S1 to S3, obtained via SPAR-H, are also presented in Table~\ref{Ps}. 

$P_{t}$ uses the time available ($T_{avail}$) and time required ($T_{reqd}$) to perform an action. To calculate $P_{t}$, $T_{reqd}$ is represented by its cumulative distribution function $F_{T_{reqd}}(t))$, and $T_{avail}$ is represented by its probability density function $F_{T_{avail}}(t)$, and $P_{t}$ is estimated as the convolution of the two probability distributions
\begin{equation} \label{Pt}
    P_{t}=P(T_{reqd}>T_{avail}) = \int_{0}^{\infty } (1-F_{T_{reqd}}(t))\cdot F_{T_{avail}}(t)dt
\end{equation}

In the IDHEAS-ECA framework, the distributions of $T_{reqd}$  and $T_{avail}$ are key components in estimating time-related failure probabilities. To model these distributions, the method typically uses two empirical percentiles (e.g., the 5th and 95th percentiles) to infer the parameters of candidate probability distributions. Commonly adopted distribution types include gamma, Weibull, lognormal, and normal distributions.

\begin{figure}[h]
\centering
\includegraphics[width=0.8\textwidth]{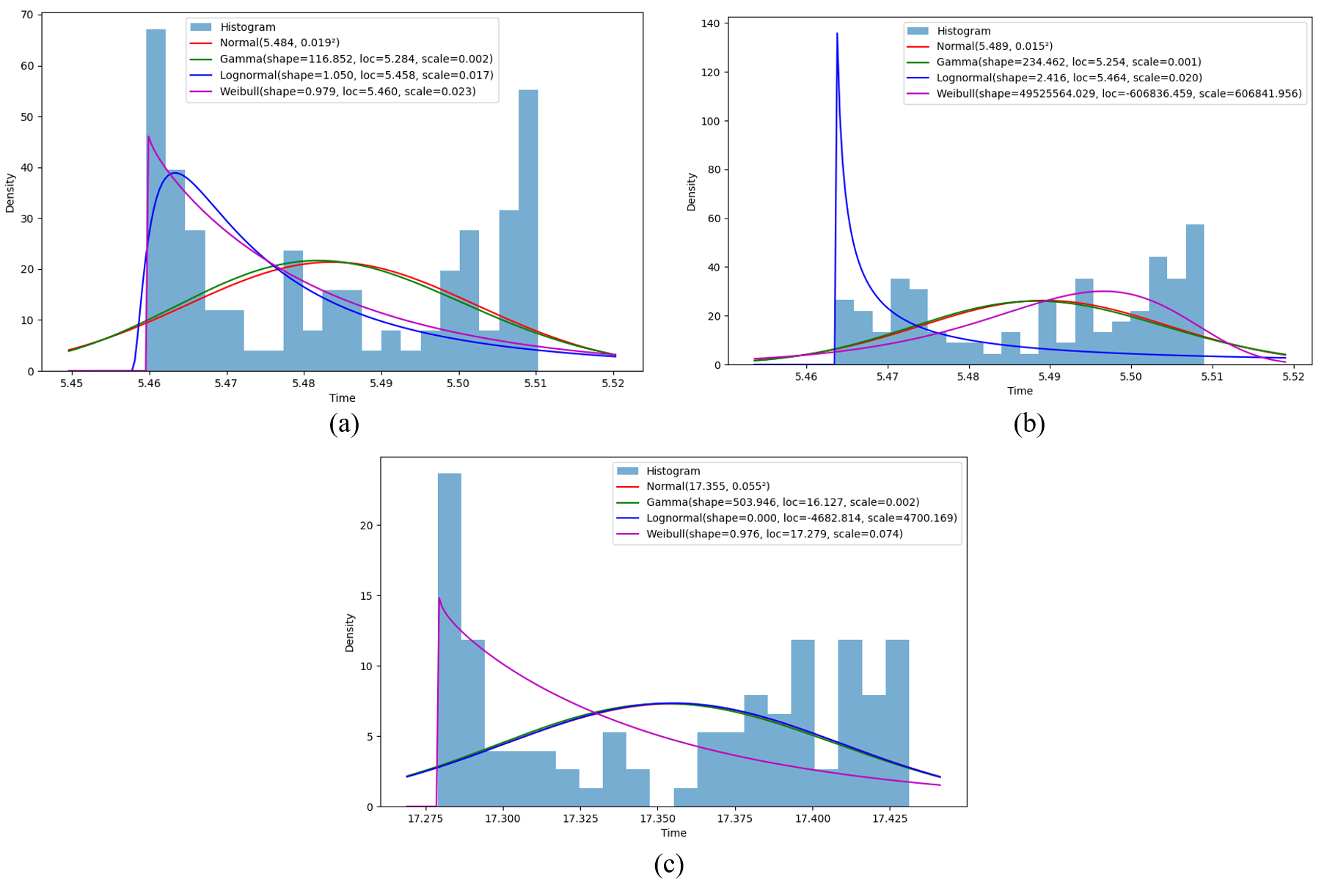}
\caption{Fitting Distributions to Hybrid-Generated Task Duration Data of Exp1}\label{distribution}
\end{figure}

In this paper, we used synthetic time-series data generated by our hybrid framework, specifically focusing on Steps 1 to 3 of Exp1. We conducted distribution fitting using the four commonly used families mentioned above. The fitting results are illustrated in Figure~\ref{distribution}, where subfigures (a), (b), and (c) respectively show the goodness-of-fit for the gamma, Weibull, lognormal, and normal distributions for each step.

The range of the time available is estimated. The lower and upper bounds of the estimate are assumed to be the 2.5th and 97.5th percentiles of any assumed probability distribution. Assuming that the time available is lognormally distributed, the results are $\mu$ = 3.50 and $\sigma $ = 0.5 for S1, $\mu$ = 6 and $\sigma $ = 0.15 for S2, and $\mu$ = 2.99 and $\sigma $ = 0.15 for S3.

\begin{table}[h]
\caption{Quantitative Comparison Between TimeGAN-Generated and Real Data human error probability}\label{Ps}%
\begin{tabular}{ p{0.15\linewidth} p{0.1\linewidth} p{0.1\linewidth} p{0.12\linewidth} p{0.12\linewidth}}
\toprule
\textbf{Procedure} & \textbf{$P_{c}$} & \textbf{$P_{t}$}  & $P_{HFE}$ & $P_{SPAR-H}$\\
\midrule
E-0 & 8.20e-3&  0.0005 & 8.70e-03 & 3.87e-3  \\
E-1 & 6.19e-3&  0.0001  &6.29e-3 & 1.38e-3\\
ES-1.2 & 8.20e-3&  0.0005 & 8.70e-3 &3.87e-3 \\
\bottomrule
\end{tabular}
\end{table}

% 此外，我们还对比了KRAIL方法，KRAIL方法是Xiao et al. 在2024年提出的\cite{xiao2024krail}. 他们基于IDHEAS-DATA，和大语言模型+知识图谱，实现了HEP的自动化分析。KRAIL方法虽然能够自动化分析huamn error rate, 但是，有个不好的点，就是数据仅限于IDHEAS-DATA所囊括的类型，对于超出数据库的数据，无法分析。

% According to the dynamic risk-informed framework (DRIF), the relationship between the overall human error probability (HEP) in exp1 and the HEPs of its constituent sub-steps can be expressed as shown in Equation~\ref{HEP}. The overall HEP is not merely a static aggregation, but rather a function that accounts for the propagation of uncertainty and behavioral dependencies across steps. Specifically, the HEP of each subsequent task is influenced by the outcomes and HEP distribution of preceding tasks.

% \begin{equation}\label{HEP}
%     HEP = 1-(1-{P_{E0}})(1-{P_{E1}})(1-{P_{ES1.2}})
% \end{equation}
% where, $P_{E0},P_{E1},P_{ES1.2}$ represent the HEPs associated with the execution of procedures E-0, E-1, and ES-1.2, respectively. As a result, the final HEP estimated using the IDHEAS-ECA method was $2.35×10^{-2}$, whereas the HEP calculated using the SPAR-H method was $9.09×10^{-3}$.

Finally, a sensitivity analysis was conducted to assess the impact of different distributional assumptions for $T_{reqd}$  on the estimated HEPs. Specifically, we evaluated how the choice of fitted distribution affects the resulting $T_{reqd}$ and overall HEP ($P_{c}$) values for each step in Exp1. The sensitivity analysis results for S1 are presented in Table~\ref{Sensitivity1}. Four candidate distributions, lognormal, gamma, Weibull, and normal, were fitted to the TimeGAN-generated task duration data for S1. Despite the different parameterizations across the distributions, the resulting values of $P_{t}$ and HEP remain consistent. This indicates a high degree of robustness in the HEP estimation process under varying assumptions of the $T_{reqd}$ distribution, at least for S1.

\begin{table}[h]
\caption{Sensitivity Analysis Results of S1}\label{Sensitivity1}%
\begin{tabular}{p{0.6 \linewidth}p{0.1\linewidth} p{0.1\linewidth}}
\toprule
\textbf{$T_{reqd} Distribution Parameters$} &  \textbf{$P_{t}$}  & $HEP$ \\
\midrule
Lognormal (shape=1.050, loc=5.458, scale=0.017)&  0.0005 & 8.70e-3  \\
Gamma (shape=116.852, loc=5.284, scale=0.002)&  0.0005 &  8.70e-3  \\
Weibull (shape=0.979, loc=5.460, scale=0.023) & 0.0005&   8.70e-3  \\
Normal (5.484,$0.019^{2}$) &0.0005 &  8.70e-3 \\
\bottomrule
\end{tabular}
\end{table}

For S2, although the Weibull distribution was initially considered, it exhibited extreme parameter values and poor fitting performance, rendering it unsuitable for reliable modeling in this context. Therefore, the analysis was limited to the three better-fitting distributions: lognormal, gamma, and normal. The results of this analysis are summarized in Table~\ref{Sensitivity2}. Across all three distributional assumptions, the resulting values of $P_{t}$ and HEP remain consistently low. Specifically, the estimated HEP ranges from $6.19×10^{-3}$ to $6.29×10^{-3}$, while $P_{t}$ values vary between 0.0000 and 0.0001. These findings suggest that the overall HEP estimation for S2 is not highly sensitive to the specific choice among the acceptable distribution types, reinforcing the robustness of the approach under different statistical assumptions.

\begin{table}[h]
\caption{Sensitivity Analysis Results of S2}\label{Sensitivity2}%
\begin{tabular}{p{0.6 \linewidth}  p{0.1\linewidth} p{0.1\linewidth}}
\toprule
 \textbf{$T_{reqd} Distribution Parameters$} & \textbf{$P_{t}$}  & $HEP$ \\
\midrule
 Lognormal (shape=2.416, loc=5.464, scale=0.020) & 0.0001 & 6.29e-3 \\
Gamma (shape=234.462, loc=5.254, scale=0.001)&  0.0000 & 6.19e-3  \\
Normal (5.489,$0.015^{2}$) &  0.0000 &  6.19e-3 \\
\bottomrule
\end{tabular}
\end{table}

For S3, both the lognormal and gamma distributions were found to be inappropriate, as they exhibited poor goodness-of-fit and produced unrealistic parameter estimates. As a result, the analysis was restricted to the two distributions that demonstrated reasonable fit: Weibull and normal. The sensitivity analysis results are presented in Table~\ref{Sensitivity3}. Despite differences in the parameterization of the two distributions, both yield the same value of $P_{t}$, and consequently, an identical overall HEP of $3.87×10^{-3}$. This consistency indicates that, for S3, the HEP estimation is robust to distributional assumptions when restricted to suitably fitted models.

\begin{table}[h]
\caption{Sensitivity Analysis Results of S3}\label{Sensitivity3}%
\begin{tabular}{p{0.6 \linewidth}  p{0.1\linewidth} p{0.1\linewidth}}
\toprule
 \textbf{$T_{reqd} Distribution Parameters$} & \textbf{$P_{t}$}  & $HEP$ \\
\midrule
Weibull (shape=0.976, loc=17.279, scale=0.074) &0.0005 & 3.87e-3 \\
Normal (17.355,$0.055^{2}$) & 0.0005 & 3.87e-3   \\
\bottomrule
\end{tabular}
\end{table}

\subsection{Influence Strength and Sensitivity Analysis based on Bayesian Network}

Drawing on the human factors risk assessment module within the dynamic risk-informed framework (DRIF), we employed monte carlo (MC) simulation to generate probabilistic data. Based on this data and our previously developed procedural framework for HRA data generation, we constructed a Bayesian Network to perform influence strength assessment and sensitivity analysis.

Unlike the traditional DRIF \cite{xiao2025dynamic} implementation, however, the HEP associated with each node in our framework is not based on expert judgment or static estimates. Instead, each node's HEP is quantitatively derived using our proposed hybrid framework, which integrates TimeGAN and ACT-R simulation, and the IDHEAS-ECA methodology. This allows for a more data-driven and context-sensitive estimation of human error probabilities, capturing both task-specific dynamics and cognitive performance factors systematically.

\begin{figure}[h]
\centering
\includegraphics[width=0.8\textwidth]{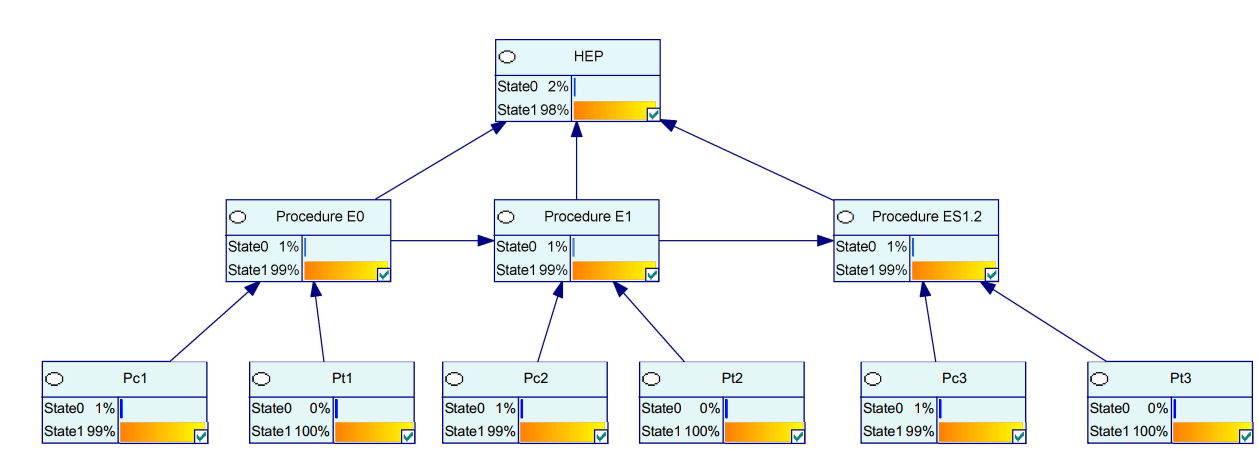}
\caption{The Bayesian Network Constructed in the Proposed Framework COGMIF}\label{bayesian}
\end{figure}

In this paper, each node in the Bayesian network corresponds to a specific procedural step and is modeled with two possible states: S1 (Failure) and S2 (Success). The relationship between the overall Human Error Probability (HEP) in Experiment 1 (Exp1) and the HEPs of its constituent sub-steps is represented in Figure~\ref{bayesian}. The Bayesian network was constructed using GeNIe 2.0, a widely used tool for probabilistic graphical modeling and decision analysis.

\begin{figure}[h]
\centering
\includegraphics[width=0.8\textwidth]{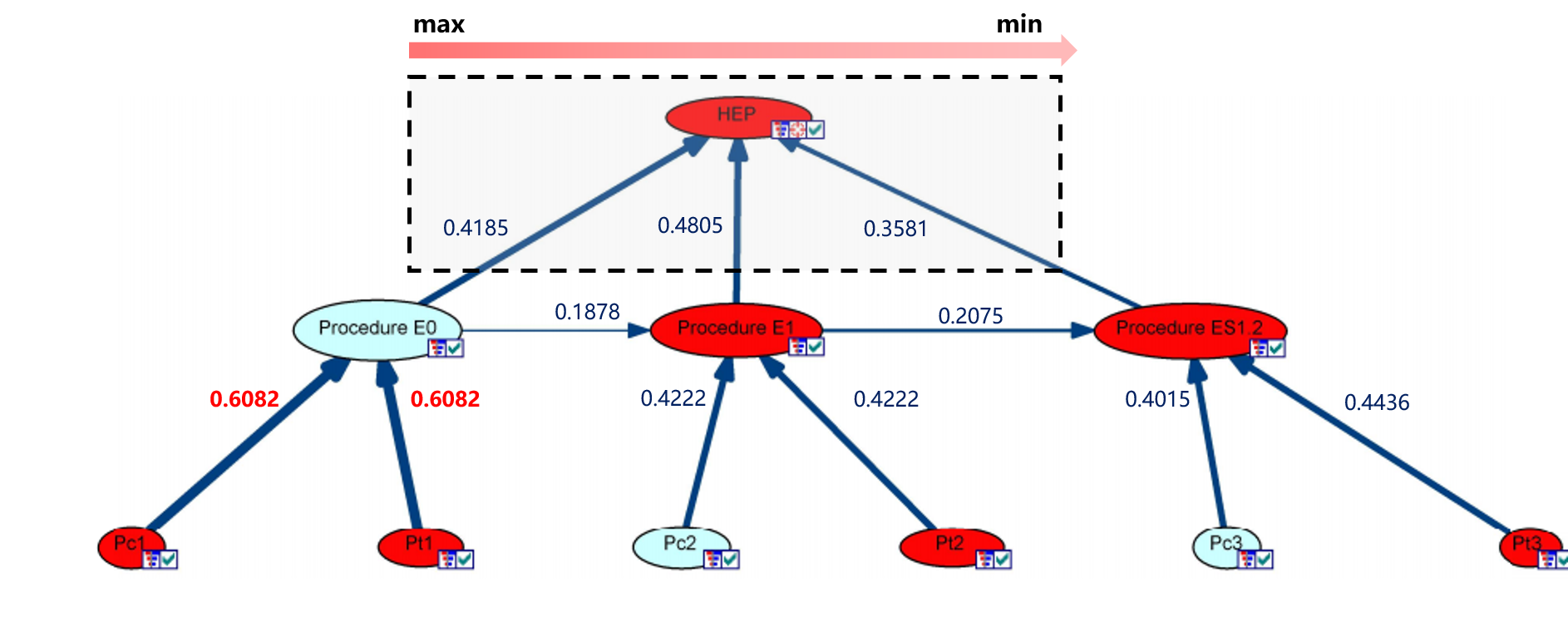}
\caption{Influence Strength Analysis in COGMIF on Overall Human Error Probability in Exp1}\label{influence}
\end{figure}

As shown in the Figure~\ref{influence}, the relative importance of procedural steps E-0, E-1, and ES-1.2 on the overall HEP decreases progressively, indicating that earlier steps in the procedure have a greater impact on the final risk outcome. Furthermore, within the execution of Procedure E-0, the cognitive error probability $P_{c1}$  and the time-related failure probability $P_{t1}$  exhibit particularly high importance scores, suggesting that both cognitive and time pressure factors play a dominant role in shaping the likelihood of failure at this critical initial stage.

Next, to observe the degree to which each variable impacts the overall risk, we conducted a sensitivity analysis employing the maximum approach. It is typically aimed at observing the system’s response under extreme conditions.

Table~\ref{Sensitivity} presents the results of a sensitivity analysis assessing the impact of both procedural steps and cognitive variables on the overall Human Error Probability (HEP) in Experiment 1. The analysis was performed using the Bayesian network constructed under the proposed hybrid framework, quantifying the relative influence of each variable on the final outcome. As shown in Table \ref{Sensitivity}, the procedural nodes, particularly Procedure ES-1.2 and Procedure E-1, demonstrate the highest sensitivity values, indicating their dominant contribution to variations in overall HEP. Among the cognitive and time-related variables, the time-related failure probabilities 
$P_{t1}$, $P_{t2}$, and $P_{t3}$ exhibit notably high sensitivity, especially $P_{t1}$, which corresponds to the execution of Procedure E-0. This highlights the critical role of time constraints in influencing human reliability during early procedural stages.

Cognitive error probabilities $P_{c1}$ , $P_{c2}$, and $P_{c3}$   also contribute to the overall HEP, albeit with relatively lower sensitivity values compared to their time-related counterparts. Interestingly, while Procedure E-0 is operationally important, its direct sensitivity score is comparatively low, suggesting that its influence on overall risk may be more indirect, mediated through cognitive and temporal subcomponents. These results provide valuable insights into which procedural elements and human factors warrant prioritization in future risk mitigation strategies and interface design improvements.

\begin{table}[h]
\caption{Sensitivity Analysis in COGMIF on Overall Human Error Probability in Exp1}\label{Sensitivity}%
\begin{tabular}{p{0.1 \linewidth}  p{0.2\linewidth} p{0.2\linewidth}}
\toprule
 \textbf{Item} & \textbf{Variable}  & Sensitivity \\
\midrule
(1) & Procedure ES1.2 & 6.24e5 \\
(2) & Procedure E1 & 5.49e5 \\
(3) & $P_{t1}$ & 920 \\
(4) &$P_{t3}$& 880 \\
(5) &$P_{t2}$& 879 \\
(6) &$P_{c1}$& 123 \\
(7) &$P_{c2}$& 98.1 \\
(8) &$P_{c3}$& 97.9 \\
(9) &Procedure E0& 69.9 \\

\bottomrule
\end{tabular}
\end{table}

\section{Conclusion and Discussion} \label{Conclusion and Discussion}

This study presents a cognitive-mechanistic human reliability analysis framework (COGMIF) that enhances the IDHEAS-ECA methodology through the integration of an ACT-R-based human digital twin (HDT) and a GAN-augmented behavioral data generation approach. Unlike traditional HRA data collection methods that rely heavily on expert judgment and resource-intensive human-in-the-loop simulations, the proposed framework leverages the ACT-R cognitive architecture, incorporating declarative memory, procedural rules, and perceptual-motor mechanisms, to simulate the cognitive and behavioral dynamics of nuclear power plant operators under a variety of operational scenarios.

The framework COGMIF was rigorously evaluated using a high-temperature gas-cooled reactor (HTGR) simulator across representative tasks, including multi-parameter queries with interface switching, single-parameter assessments, and three-phase AC operations. Results demonstrate that the ACT-R-based HDT effectively predicts task completion times, offering a cognitively interpretable and mechanistically grounded alternative to conventional HRA models.

To overcome the scalability limitations of repeated cognitive simulations, the framework incorporates TimeGAN, a generative adversarial model trained on ACT-R-generated time-series data, which enables the synthesis of large-scale, high-fidelity behavioral datasets that preserve temporal structure and cognitive realism. These synthetic datasets are subsequently integrated into the IDHEAS-ECA framework to enable structured, mechanism-informed estimation of HEPs consistent with modern probabilistic risk assessment (PRA) standards.

Furthermore, the hybrid-generated data were used to compute the time-based failure probability ($P_t$) component of IDHEAS-ECA, offering a data-driven alternative to expert-derived duration thresholds. To support higher-level causal analysis, cognitive and procedural features were mapped onto a Bayesian network, facilitating influence strength quantification and sensitivity analyses to identify key contributors to overall human error.

Sensitivity studies in COGMIF confirmed the robustness of the proposed approach under varying distributional assumptions, while comparative evaluations with the SPAR-H method further validated the model’s reliability and its ability to reflect scenario-specific cognitive dynamics.

Collectively, this work contributes a validated, scalable, and cognitively interpretable HRA framework that addresses both the methodological gaps and data constraints of existing models. It offers a credible pathway for enhancing human reliability modeling in support of advanced nuclear system design, safety assessment, and operator performance evaluation in next-generation control environments.

Looking ahead, future research should incorporate advanced simulation environments, large language models (LLMs) \cite{chen2025combatvla}, and 3D digital human representations \cite{chen2023diffusiontalker, chen2024mixedgaussianavatar} holds significant promise for enhancing operator training and optimizing reactor safety protocols.

\section*{Declarations}

\subsection*{Conflict of interest/Competing interests} 
The authors declare that they have no known competing financial interests.
\subsection*{Funding}
The research was supported by a grant from the National Natural Science Foundation of China (Grant No. T2192933) and the Foundation of National Key Laboratory of Human Factors Engineering (Grant NO.HFNKL2024W07).
\subsection*{Author contribution} Xingyu Xiao: Conceptualization, Methodology, Software, Formal analysis, Data Curation, Visualization, Validation, Writing- Original draft preparation. Peng Chen: Software, Methodology. Jiejuan Tong: Conceptualization, Formal analysis, Supervision, Writing - Review and Editing. Shunshun Liu: Methodology. Hongru Zhao: Supervision, Writing - Review and Editing. Jun Zhao: Supervision, Writing - Review and Editing. Qianqian Jia: Supervision, Writing - Review and Editing. Jingang Liang: Resources, Supervision, Writing - Review and Editing, Project administration, Funding acquisition. Wang Haitao: Supervision, Writing- Reviewing and Editing.

\noindent

\bibliography{sn-bibliography}% common bib file

\end{document}